\documentclass[conference]{IEEEtran}

\IEEEoverridecommandlockouts

\usepackage{amsmath}
\usepackage{enumitem}
\usepackage{booktabs}
\usepackage{graphicx}
\usepackage{subcaption}
\usepackage{xcolor}
\usepackage{adjustbox}
\usepackage{xspace}
\usepackage{algorithm}
\usepackage{algpseudocode}
\usepackage{multirow}
\usepackage[numbers]{natbib}
\usepackage{amsthm}
\usepackage{amsfonts}
\usepackage[hidelinks]{hyperref}
\newtheorem{example}{Example}
\algtext*{EndIf}
\algtext*{EndFor}
\algtext*{EndWhile}
\usepackage[T1]{fontenc}
\usepackage[utf8]{inputenc}

\newcommand{\ignore}[1]{}

\newcommand{\nth}{\textit{nth}\xspace}
\newcommand{\cstart}{\text{cleanstart}}

\newcommand{\cifar}{\texttt{CIFAR10}\xspace}
\newcommand{\cifarh}{\texttt{CIFAR10-H}\xspace}

\newcommand{\mnist}{\texttt{MNIST}\xspace}
\newcommand{\adult}{\texttt{Adult}\xspace}

\newcommand{\cilnbench}{\textsc{CILN}\xspace}
\newcommand{\cilnbenchS}{\textsc{CILN-S}\xspace}
\newcommand{\cilnbenchC}{\textsc{CILN-C}\xspace}
\newcommand{\plidn}{\textsc{PL-IDN}\xspace}
\newcommand{\plidnL}{\textsc{PL-IDN-L}\xspace}
\newcommand{\plidnM}{\textsc{PL-IDN-M}\xspace}
\newcommand{\plidnH}{\textsc{PL-IDN-H}\xspace}

\newcommand{\bright}{\texttt{bright}}
\newcommand{\contr}{\texttt{contr}}

\newcommand{\fog}{\texttt{fog}}
\newcommand{\frost}{\texttt{frost}}
\newcommand{\snow}{\texttt{snow}}

\newcommand{\jpeg}{\texttt{jpeg}}

\newcommand{\rotate}{\texttt{rotate}}

\newcommand{\shear}{\texttt{shear}}

\newcommand{\scaling}{\texttt{scaling}}

\newcommand{\ctype}[1]{\texttt{#1}}
\newcommand{\cfamily}[1]{\texttt{#1}}

\title{[Experiment, Analysis, and Benchmark] Benchmarking Instance-Dependent Label Noise with Controlled Corruptions}

\author{
\IEEEauthorblockN{Shadman Islam, Agustinus Kristiadi, Mostafa Milani}
\IEEEauthorblockA{
Department of Computer Science, Western University\\
London, Ontario, Canada\\
\{misla2, akristi, mostafa.milani\}@uwo.ca
}
}

\begin{document}

\maketitle

\begin{abstract}
Synthetic instance-dependent label noise (IDN) benchmarks are widely used to evaluate noisy-label learning methods, yet existing approaches typically generate noise through imperfect annotators or classifier raters, leaving the source of ambiguity implicit. We introduce \cilnbench{}, a benchmark-generation framework that creates IDN through controlled input corruptions. A diverse voter pool labels corrupted instances, producing benchmark datasets in which both the source and severity of ambiguity are explicit and controllable. Using CIFAR-10, MNIST, and Adult, we construct 90 benchmark settings spanning multiple corruption families and severity levels. Our experiments show that the resulting benchmarks exhibit genuine instance-dependent noise, provide diverse confusion structures, and, on CIFAR-10, can produce label distributions that are closer to human uncertainty than an existing synthetic IDN benchmark. We further demonstrate that corruption-mediated IDN can expose failure modes of popular noisy-label learning methods, including Co-Teaching and DivideMix, that are not observed under comparable levels of rater-fallibility noise. These findings suggest that noise structure, not only noise rate, plays an important role in benchmark difficulty and algorithm behavior. By making ambiguity generation explicit and controllable, \cilnbench{} provides a complementary benchmarking framework for studying noisy-label learning under diverse sources of instance difficulty.
\end{abstract}


\section{Introduction}

Modern machine learning systems rely on large labeled datasets, yet labels are often imperfect due to annotation mistakes, ambiguous examples, weak supervision pipelines, or automatically generated annotations~\cite{natarajan2013learning,patrini2017loss,han2018coteaching}. Such label noise can substantially degrade downstream performance by reducing predictive accuracy and introducing bias~\cite{rolnick2017robust,jiang2020mentormix,lu2023tabasco}. While early work primarily studied \emph{symmetric} and \emph{class-conditional} noise, these assumptions often fail in practice because labeling errors depend not only on the class but also on the particular example being labeled. Difficult or ambiguous instances are naturally more likely to receive incorrect labels than easy ones. This more realistic setting, in which corruption probabilities depend on both the class and the individual instance, is commonly referred to as \emph{instance-dependent label noise} (IDN)~\cite{xia2020partdependent,berthon2021confidence}.

Despite its practical importance, evaluating methods under IDN remains challenging due to the limited availability of realistic and controllable benchmarks. Existing approaches generally fall into two categories. Early synthetic methods generate IDN through instance-specific corruption probabilities or feature-space transformations~\cite{xia2020partdependent,berthon2021confidence}, making some examples more likely than others to receive incorrect labels. More recently, pseudo-labeling approaches generate IDN using diverse classifiers acting as synthetic annotators, where disagreement among model predictions induces label uncertainty~\cite{gu2022idn}. These approaches provide practical mechanisms for generating IDN and can produce noise levels and uncertainty patterns that resemble human annotation behavior.

While these approaches provide practical mechanisms for generating IDN, the source of uncertainty remains largely implicit. In feature-space approaches, label errors arise from latent transformations that are often difficult to interpret. In pseudo-labeling approaches, uncertainty emerges through disagreement among synthetic annotators, and noise levels are typically controlled by modifying properties of the annotators themselves, such as their training procedure or predictive quality. As a result, although these benchmarks can generate realistic instance-dependent noise, the ambiguity responsible for a particular label error is often difficult to identify or manipulate directly. Consequently, existing benchmarks provide limited support for systematically studying how specific ambiguity mechanisms---such as blur, occlusion, missing information, or measurement errors---affect classifier disagreement, label uncertainty, and downstream learning behavior.

In this paper, we introduce \cilnbench\ (\emph{Corruption-Induced Label Noise}), a framework for generating instance-dependent label-noise benchmarks through controlled input corruptions. A \emph{corruption} is a transformation that degrades or alters the original input while preserving its semantic content, such as blur, noise, geometric distortions, missing values, or attribute perturbations. Starting from clean instances, \cilnbench\ applies controlled corruptions at varying severity levels~\cite{hendrycks2019corruptions,mu2021mnistc,jenga2023} and measures their effect on the predictions of a diverse voter pool. For image datasets, we consider corruption families such as noise, blur, weather, and geometric transformations, while for tabular datasets we consider missing-data and value corruptions. The corrupted instances are then evaluated by the voter pool to generate noisy labels and soft label distributions. As a result, label uncertainty emerges from observable and explicitly controlled ambiguity mechanisms whose type and severity are known.

Unlike existing IDN benchmarks, which primarily vary the labeling process to generate noisy labels, \cilnbench\ generates label uncertainty by modifying the input itself through controlled corruptions. This shifts the focus from controlling how noisy labels are produced to controlling why they arise in the first place. As a result, \cilnbench\ enables systematic studies of how specific corruption mechanisms affect classifier disagreement, label uncertainty, and downstream learning behavior. Moreover, because the type and severity of each corruption are known, the resulting label transitions can often be traced to specific changes in the input. The following example illustrates this property.

\begin{example}
\label{ex:severity_arc}

Figure~\ref{fig:severity_arc} shows four representative \cilnbench\ examples---two from \cifar\ and two from \mnist\---under increasing corruption severity. Each column follows a single example from mild to severe corruption for a fixed corruption type. The examples illustrate how controlled corruptions can progressively increase classification difficulty and ultimately change the most plausible label.

\begin{itemize}[leftmargin=1em]
\item \textbf{(a) \cifar, \ctype{contrast}, true: dog.}
As contrast drops, the fine fur, snout, and ear details that distinguish the animal as a dog gradually disappear, leaving primarily its silhouette. Because cats and dogs share similar body shapes, the corrupted image increasingly resembles a black cat, leading to a plausible dog$\rightarrow$cat transition.
\item \textbf{(b) \cifar, \ctype{gaussian}, true: bird.}
The original image already contains visual cues compatible with multiple classes. As Gaussian noise increases, the bird becomes harder to distinguish from the noisy background, and the resulting appearance increasingly resembles a frog photographed in vegetation, producing a bird$\rightarrow$frog transition.
\item \textbf{(c) \mnist, \ctype{impulse}, true: 5.}
Impulse noise progressively fills gaps in the digit. At moderate severity, the lower curve begins to resemble part of an 8, while at higher severity both open regions become closed. The resulting digit is visually consistent with an 8, leading to a 5$\rightarrow$8 transition.
\item \textbf{(d) \mnist, \ctype{shot}, true: 4.}
Shot noise thickens the strokes of the digit and gradually fills the open region at the top of the 4. As the rectangular opening closes and becomes more loop-like, the digit increasingly resembles a 9, leading to a plausible 4$\rightarrow$9 transition.
\end{itemize}

\begin{figure}[t]
\centering

\begin{subfigure}[t]{0.24\linewidth}
\centering
\includegraphics[width=\linewidth]{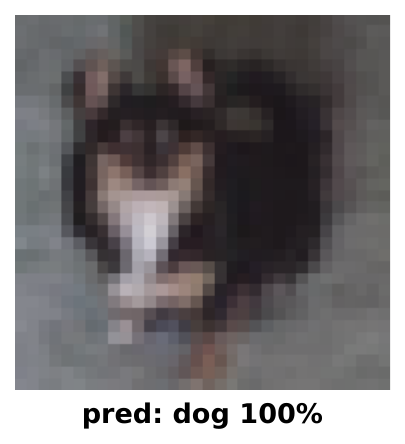}\\[-0.1em]
\includegraphics[width=\linewidth]{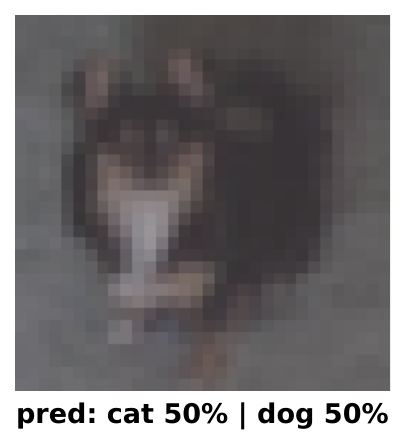}\\[-0.1em]
\includegraphics[width=\linewidth]{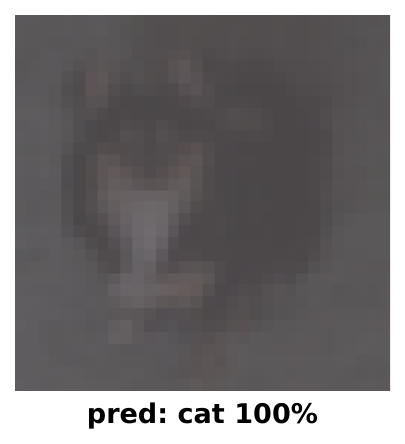}
\caption{\ctype{contrast}}
\label{fig:sev_a}
\end{subfigure}
\hfill
\begin{subfigure}[t]{0.24\linewidth}
\centering
\includegraphics[width=\linewidth]{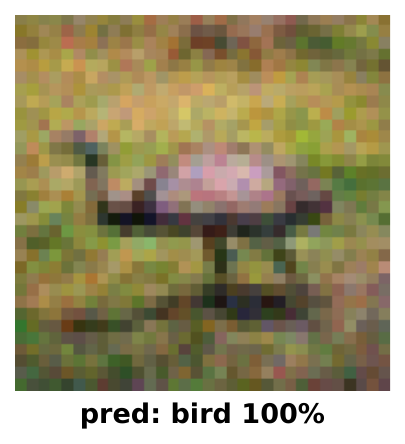}\\[-0.1em]
\includegraphics[width=\linewidth]{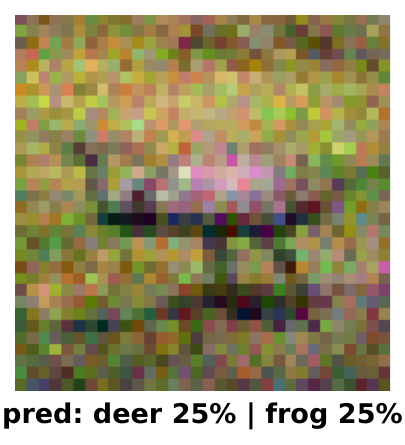}\\[-0.1em]
\includegraphics[width=\linewidth]{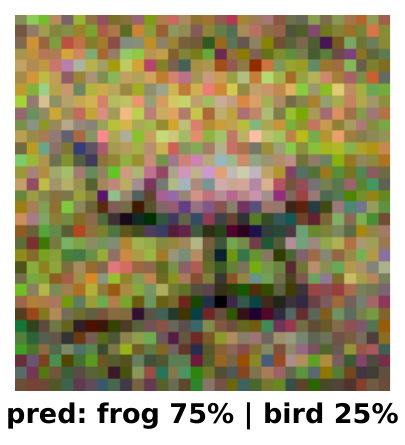}
\caption{\ctype{gaussian}}
\label{fig:sev_b}
\end{subfigure}
\hfill
\begin{subfigure}[t]{0.24\linewidth}
\centering
\includegraphics[width=\linewidth]{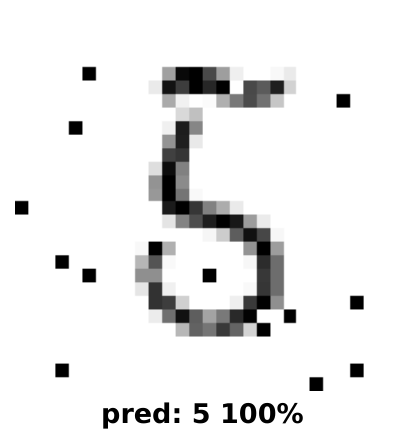}\\[-0.1em]
\includegraphics[width=\linewidth]{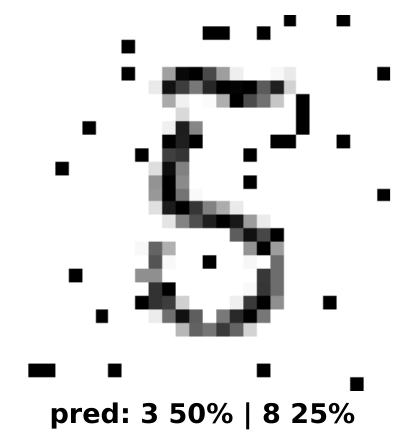}\\[-0.1em]
\includegraphics[width=\linewidth]{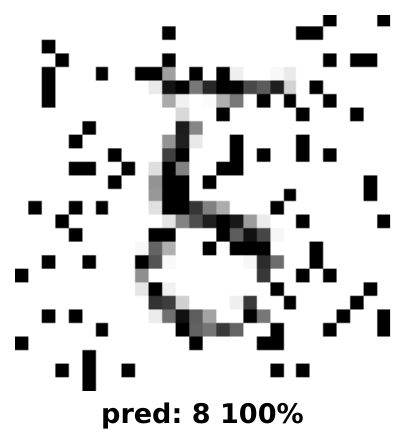}
\caption{\ctype{impulse}}
\label{fig:sev_c}
\end{subfigure}
\hfill
\begin{subfigure}[t]{0.24\linewidth}
\centering
\includegraphics[width=\linewidth]{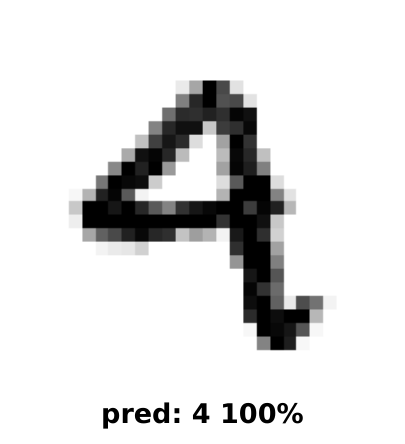}\\[-0.1em]
\includegraphics[width=\linewidth]{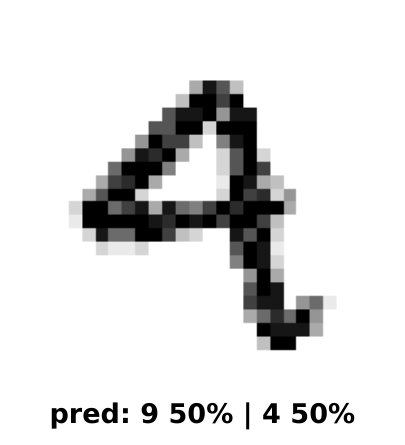}\\[-0.1em]
\includegraphics[width=\linewidth]{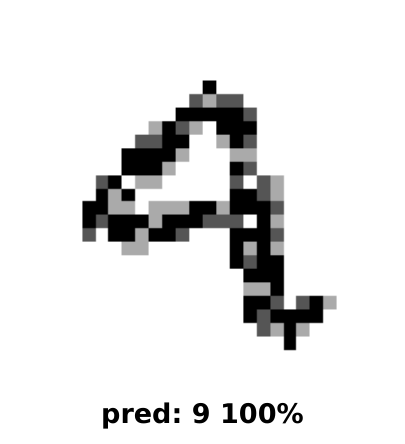}
\caption{\ctype{shot}}
\label{fig:sev_d}
\end{subfigure}

\caption{Examples of corruption-induced label ambiguity under increasing corruption severity (top to bottom).}
\label{fig:severity_arc}
\end{figure}

These examples illustrate that corruption-induced label transitions arise from identifiable changes in the input rather than arbitrary label flips. As corruption severity increases, evidence supporting the true class gradually weakens while features associated with alternative classes become more prominent. Consequently, the resulting label uncertainty can often be traced to a specific corruption mechanism, providing a transparent and controllable source of instance-dependent label noise.\end{example}

The explicit control provided by \cilnbench\ enables several forms of analysis that are difficult or impossible to perform with existing IDN benchmarks. First, because the source and severity of each corruption are known, benchmark instances can be generated under specific corruption mechanisms, allowing systematic comparisons across different sources of input difficulty. Second, an optional \emph{clean-start} configuration retains only instances that are unanimously classified correctly before corruption, enabling corruption-induced label uncertainty to be isolated from pre-existing ambiguity in the original data. Finally, by varying corruption types and severity levels, \cilnbench\ produces benchmark instances with different noise characteristics and levels of instance dependence, enabling controlled studies of how these factors affect downstream learning methods.

These capabilities also distinguish \cilnbench\ from existing human-uncertainty and pseudo-label benchmarks such as CIFAR10-H~\cite{peterson2019cifar10h} and Gu et al.~\cite{gu2022idn}. Existing benchmarks primarily generate label uncertainty through human disagreement or synthetic annotators, whereas \cilnbench\ generates uncertainty through controlled modifications of the input itself. As a result, \cilnbench\ enables studies not only of how much label noise is present, but also of how specific corruption mechanisms contribute to classifier disagreement, label uncertainty, and downstream learning behavior. The two perspectives are complementary: one focuses on the labeling process, while the other focuses on the characteristics of the data that make labeling difficult.

We evaluate \cilnbench\ on CIFAR-10, MNIST, and Adult, covering both image and tabular domains. First, we demonstrate that controlled corruptions generate genuinely instance-dependent label noise and that the degree of instance dependence can be systematically influenced through corruption type and severity. Second, we show that particular corruption settings induce confusion and uncertainty patterns whose distributions are statistically closer to human annotation uncertainty, such as that observed in CIFAR10-H~\cite{peterson2019cifar10h}, than existing synthetic IDN benchmark-generation approaches. Third, we evaluate state-of-the-art noisy-label learning methods, including Co-Teaching~\cite{han2018coteaching} and DivideMix~\cite{li2020dividemix}, on the generated benchmarks and show that their behavior depends not only on the overall noise rate but also on the underlying corruption mechanism. Together, these results demonstrate that controlled corruptions provide a practical and complementary approach to constructing realistic, interpretable, and mechanism-controlled IDN benchmarks for evaluating methods under instance-dependent label noise.

Our technical contributions are summarized as follows:

\begin{itemize}[leftmargin=1em]

\item We propose \cilnbench, a corruption-mediated framework for generating mechanism-controlled IDN benchmarks. The framework introduces a clean-start condition and explicit corruption settings that establish a direct link between the source of ambiguity and the resulting label uncertainty.

\item We show that the proposed framework generates genuine IDN and that the degree of instance dependence can be systematically controlled by the type and severity of corruption.

\item We demonstrate that controlled corruptions induce diverse confusion and uncertainty patterns and that, under appropriate corruption settings, the resulting label distributions are statistically closer to human annotation uncertainty than existing synthetic IDN benchmark-generation approaches.

\item We evaluate state-of-the-art noisy-label learning methods on the generated benchmarks and show that their performance depends not only on the overall noise rate but also on the underlying ambiguity mechanism, highlighting benchmark characteristics that are not captured by traditional synthetic noise models.

\end{itemize}

\section{Preliminaries}

This section introduces the notation and concepts used in the paper. Section~\ref{sec:prelim-noise} reviews the classification setting and label-noise models, Section~\ref{sec:prelim-soft} introduces voter-generated label distributions, and Section~\ref{sec:prelim-corruption} defines corruption settings.

\subsection{Classification Setting and Noise Models} \label{sec:prelim-noise}

We consider a $K$-class classification problem with input space $\mathcal{X}$ and label space $\mathcal{Y}=\{1,\ldots,K\}$. Let $D=\{(x_i,y_i)\}_{i=1}^{N}$ denote a clean dataset, where $x_i\in\mathcal{X}$ is an input sample and $y_i\in\mathcal{Y}$ is its ground-truth class label. Depending on the dataset, $x_i$ may represent an image or a tabular record.

Label noise occurs when the observed label $\tilde y_i$ differs from the clean label $y_i$. The noisy dataset is written as $
\tilde D=\{(x_i,\tilde y_i)\}_{i=1}^{N}$. Existing work commonly studies several noise-generation settings. In symmetric label noise, labels are randomly flipped with fixed probability $\delta$:%
\[
P(\tilde y=j \mid y=i)
=
\begin{cases}
1-\delta, & j=i,\\[2mm]
\dfrac{\delta}{K-1}, & j\neq i.
\end{cases}
\]

\noindent Class-conditional noise (CCN), assumes that the corruption process depends only on the true class label and not on the input instance. Formally,%
\[
P(\tilde{y}=j \mid x, y=i)
=
P(\tilde{y}=j \mid y=i).
\]

\noindent Under this assumption, the corruption process can be represented by a transition matrix $T\in\mathbb{R}^{K\times K}$:%
\[
P(\tilde{y}=j \mid y=i)
=
T_{ij},
\]

\noindent where each row of $T$ specifies the corruption probabilities for class $i$. Both symmetric label noise and CCN ignore the properties of individual examples and assume that all instances from the same class share the same corruption behavior.

In practical settings, however, the probability of label corruption often depends on the input itself. Harder examples, ambiguous inputs, or degraded observations are naturally more likely to receive incorrect labels. IDN relaxes the CCN assumption by allowing the corruption process to depend on both the input instance and the true label. Consequently, the noisy-label distribution is modeled as $P(\tilde{y}\mid x,y)$, allowing examples from the same class to have different probabilities of receiving incorrect labels depending on their individual characteristics and difficulty.

\subsection{Label Distributions and Voter Pools}
\label{sec:prelim-soft}

In many classification tasks, uncertainty cannot be adequately represented by a single class label. Instead, uncertainty can be represented as a label distribution over the $K$ classes. For an input $x$, we write such a distribution as%
\[
p(y \mid x) = (p_1,\ldots,p_K),
\qquad
p_k \geq 0 \;\; \forall k,
\qquad
\sum_{k=1}^{K} p_k = 1,
\]

\noindent where $p_k$ denotes the probability assigned to class $k$.

Label distributions arise naturally when multiple annotators or classifiers provide labels for the same instance. Let $V={f_1,\ldots,f_M}$ denote a pool of $M$ voters. Given an input $x$, each voter produces either a hard label or a predictive distribution over classes. Collectively, these predictions characterize the uncertainty associated with $x$.

The predictions of multiple voters can be aggregated into a single label distribution using an aggregation operator. When voters provide hard labels, common strategies include majority voting and empirical vote frequencies. When voters provide predictive distributions, aggregation can be performed directly in distribution space. The resulting aggregate distribution summarizes the collective uncertainty of the voter pool and provides a richer representation than a single hard label.

In \cilnbench, each voter produces a predictive distribution $p^{(m)}(y \mid x)$. For each instance, the benchmark releases an aggregated label distribution $\bar p_i(y \mid x)$ obtained by arithmetic averaging:
\begin{align}
\bar p_i(y \mid x)=\frac{1}{M}\sum_{m=1}^{M} p_i^{(m)}(y \mid x).    
\end{align}

\noindent This aggregation admits a natural probabilistic interpretation: $\bar p_i$ corresponds to the expected label distribution obtained by selecting a voter uniformly at random from the pool. In addition to $\bar p_i$, \cilnbench\ stores the individual voter distributions $\{p_i^{(m)}\}_{m=1}^{M}$, which enable alternative aggregation strategies and analyses of voter disagreement and benchmark noise characteristics.

\subsection{Corruption Families} \label{sec:prelim-corruption}

The corruption types in our benchmark are drawn from established robustness benchmarks and data-quality literature~\cite{hendrycks2019corruptions,mu2021mnistc,jenga2023}, allowing us to study how different forms of data degradation induce ambiguity and disagreement among classifiers. We consider both image and tabular corruptions to demonstrate that the framework is not restricted to a single modality.

A \emph{corruption setting} is a pair
\[
\tau=(c,\ell),
\]
where \(c\) denotes a \emph{corruption type} and \(\ell\) denotes a \emph{severity level}. Corruption types specify the mechanism of degradation, while severity levels control the strength of the degradation. Following prior robustness benchmarks~\cite{hendrycks2019corruptions,mu2021mnistc}, each corruption type is associated with a set of severity levels that control the strength of the degradation. Throughout this work, we use severity levels $\ell \in \{1,3,5\}$, corresponding to mild, moderate, and severe corruption.

\subsubsection{Image Corruptions}

For image data, we consider six corruption families commonly used in robustness benchmarks~\cite{hendrycks2019corruptions,mu2021mnistc}.

\begin{itemize}[leftmargin=1em]

\item \cfamily{Noise} corruptions perturb pixel values through random disturbances. Representative corruption types include \ctype{gaussian-noise} (additive Gaussian noise), \ctype{shot-noise} (Poisson noise caused by the discrete nature of light), \ctype{impulse-noise} (salt-and-pepper style corruption caused by sparse pixel errors), and \ctype{spatter} (random blotches that partially occlude the image).

\item \cfamily{Blur} corruptions reduce spatial detail and edge information. Representative corruption types include \ctype{defocus-blur} (out-of-focus lens blur), \ctype{glass-blur} (local pixel shuffling resembling frosted glass), \ctype{motion-blur} (blur caused by camera or object motion), and \ctype{zoom-blur} (blur induced by rapid zooming).

\item \cfamily{Geometric} corruptions alter the spatial structure of the image. Representative corruption types include \ctype{elastic} (local elastic deformation of image regions), \ctype{rotation} (in-plane rotation), \ctype{shear} (slanting the image along one axis), \ctype{translation} (shifting the image spatially), and \ctype{scaling} (resizing the image content).

\item \cfamily{Weather} corruptions simulate environmental effects. Representative corruption types include \ctype{fog} (haze that reduces visibility), \ctype{frost} (ice-crystal artifacts on the image), and \ctype{snow} (snow-like visual occlusion).

\item \cfamily{Digital} corruptions mimic acquisition and post-processing artifacts. Representative corruption types include \ctype{brightness} (global intensity shifts), \ctype{contrast} (changes in intensity separation), \ctype{jpeg} (lossy compression artifacts), and \ctype{pixelate} (loss of resolution through blocky upsampling).

\item \cfamily{Structural} corruptions introduce structured overlays or shape distortions. Representative corruption types include \ctype{canny-edges} (edge-map transformation using the Canny detector), \ctype{dotted-line} (superimposed dotted strokes), \ctype{stripe} (a vertical stripe overlay), and \ctype{zigzag} (superimposed zigzag strokes).

\end{itemize}

These corruption families span a diverse range of visual degradations, from low-level pixel perturbations to large-scale structural transformations.

\subsubsection{Tabular Corruptions}

For tabular data, we adopt corruption operators inspired by the Jenga benchmark~\cite{jenga2023}. Unlike image corruptions, tabular corruptions target the quality and completeness of attributes. We consider two broad families:

\begin{itemize}[leftmargin=1em]

\item \cfamily{Missing} corruptions remove information by replacing observed values with NaN. We consider three standard missingness mechanisms: Missing Completely At Random (\ctype{mcar}), where values are removed independently with a fixed probability; Missing At Random (\ctype{mar}), where the missingness probability depends on other observed attributes; and Missing Not At Random (\ctype{mnar}), where the missingness probability depends on the value being removed.

\item \cfamily{Value} corruptions alter observed values through perturbations such as noise injection, scaling errors, unit inconsistencies, swapped values, and other feature-level distortions.

\end{itemize}

These corruptions reflect common data-quality issues in real-world tabular datasets. The exact corruption types and severity settings used for each benchmark instantiation are described in Section~\ref{sec:datasets}.

\section{Benchmark Construction}
\label{sec:pipeline}

Our goal is to generate datasets in which label uncertainty arises from controlled and observable ambiguity mechanisms. \cilnbench\ achieves this by combining controlled input corruptions with a pool of voters. Section~\ref{sec:framework} presents the general generation procedure, while Section~\ref{sec:datasets} describes dataset-specific instantiations.

\subsection{Benchmark Generation Algorithm}
\label{sec:framework}

Let $D=\{(x_i,y_i)\}_{i=1}^{N}$ denote a clean labeled dataset, $\tau=(c,\ell)$ a corruption setting, and $V=\{f_1,\ldots,f_M\}$ the voter pool. For a fixed corruption setting $\tau$, \cilnbench\ produces a benchmark instance
\begin{align}
B_\tau
=
\left\{
\bigl(x_i,\tilde{x}_i,y_i,\bar{p}_i,\ignore{\{p_i^{(m)}\}_{m=1}^{M},}\tau\bigr)
\right\}_{i=1}^{N},    \label{eq:benchmark}
\end{align}

\noindent where $\tilde{x}_i$ denotes the corrupted input, and $\bar{p}_i$ denotes the aggregated voter distribution. The benchmark-generation procedure is summarized in Algorithm~\ref{alg:cilnbench}.

\begin{algorithm}[t]
\caption{\cilnbench Benchmark Generation.}
\label{alg:cilnbench}
\begin{algorithmic}[1]
\Statex \textbf{Input:}
Clean dataset $D=\{(x_i,y_i)\}_{i=1}^{N}$,
corruption setting $\tau=(c,\ell)$,
voter pool $V=\{f_1,\ldots,f_M\}$, $\cstart \in \{\text{false},\text{true}\}$

\Statex \textbf{Output:}
Benchmark instance $B_\tau$
\State $\mathcal{I} \gets \{1,...,N\};$
\If{$\cstart$}
\For{$i \in \mathcal{I}$ \textbf{and} $m = 1,\ldots,M$}
    \State $\hat{y}^{(m)} \gets \arg\max_c f_m(x_i)[c];$
    \If{$\hat{y}^{(m)} \neq y_i$} $\mathcal{I} \gets \mathcal{I} \setminus \{i\};$ \EndIf
\EndFor
\EndIf
\State $B_\tau \gets \emptyset;$

\For{$i \in \mathcal{I}$}
    \State $\tilde{x}_i \gets C(x_i;\tau);$ \label{alg:corrupt}

    \For{$m=1,\ldots,M$}
        \State $p_i^{(m)} \gets f_m(\tilde{x}_i)$ \label{alg:vote}
    \EndFor

    \State $\bar{p}_i \gets \frac{1}{M}\sum_{m=1}^{M} p_i^{(m)};$ \label{alg:aggregate}

    \State $B_\tau \gets B_\tau \cup
    \{(x_i,\tilde{x}_i,y_i,\bar{p}_i,\tau)\};$
\EndFor
\State \Return $B_\tau$
\end{algorithmic}
\end{algorithm}

The algorithm consists of three components: a corruption operator, a voter pool, and a label aggregation mechanism, together with an optional filtering step applied before benchmark generation. When \cstart{} is enabled, the algorithm first restricts the index set $\mathcal{I}$ to examples whose clean input $x_i$ is unanimously assigned its ground-truth label $y_i$ by all voters. Consequently, any disagreement observed after corruption can be attributed to the corruption operator $C(\cdot;\tau)$ rather than pre-existing voter disagreement on the clean example. Since examples exhibiting uncertainty before corruption are removed, the resulting benchmark typically exhibits a lower aggregate noise rate and isolates corruption-induced label uncertainty. We use \cstart{} when the goal is to study the effect of a corruption setting $\tau$ in isolation; users may disable it ($\cstart=\text{false}$) when retaining the full dataset or preserving class balance is more important. 

For each retained instance, the corruption operator $C(\cdot;\tau)$ generates a corrupted version $\tilde{x}_i$ according to the specified corruption type and severity level (Line~\ref{alg:corrupt}). The corrupted instance is then evaluated by the voter pool, producing predictive distributions ${p_i^{(1)},\ldots,p_i^{(M)}}$ (Line~\ref{alg:vote}). Disagreement among these distributions captures the uncertainty induced by the applied corruption.

The voter predictions are subsequently aggregated into a single label distribution $\bar{p}_i$ (Line~\ref{alg:aggregate}), which serves as the benchmark soft label. Following Section~\ref{sec:prelim-soft}, we use arithmetic mean aggregation, so $\bar{p}_i$ corresponds to the expected label distribution obtained by selecting a voter uniformly at random from the pool.

Some downstream learning methods require a single training label per instance. For those experiments, we derive hard labels from the voter predictions using the same uniform-rater sampling protocol employed by PL-IDN \cite{gu2022idn}. Specifically, for each instance, we select one voter uniformly at random and use its prediction as the observed label.

The algorithm returns a benchmark instance $B_\tau$ consisting of tuples $(x_i,\tilde{x}_i,y_i,\bar{p}_i,\tau)$ for all retained examples. Thus, each benchmark instance contains the clean input, corrupted input, ground-truth label, aggregated label distribution, and corruption metadata. Different corruption settings $\tau$ produce different benchmark instances, while applying the algorithm to a different dataset, corruption family, or voter pool produces a new benchmark suite.

\subsection{Datasets and Benchmark Instantiations}
\label{sec:datasets}

We instantiate the algorithm of Section~\ref{sec:framework} on three datasets spanning image and tabular domains: \cifar, \mnist, and \adult. For each dataset, we construct a diverse voter pool and generate benchmark instances using corruption settings $\tau=(c,\ell)$, where corruption types $c$ are selected from the families described in Section~\ref{sec:prelim-corruption} and severity levels are fixed to $\ell\in{1,3,5}$. Table~\ref{tab:instantiations} summarizes the resulting benchmark suites.

\begin{table}[t]
\centering
\small
\begin{tabular}{lcccc}
\toprule
\textbf{Benchmark} & \textbf{Voters} & \textbf{\shortstack{Rows\\ per $\tau$}} & \textbf{\shortstack{Released\\ $\tau$}} & \textbf{\shortstack{Label noise rate\\ range}} \\
\midrule
\cifar & 4 & 22,500 & 45 & $8.3\%-75.0\%$ \\
\mnist & 4 & 27,000 & 30 & $5.9\%-71.5\%$ \\
\adult & 5 & 13,566 & 15 & $14.7\%-26.3\%$ \\
\bottomrule
\end{tabular}
\caption{Summary of the benchmark instantiations used in this work. \emph{Label noise rate ($\delta$)} denotes the average disagreement between voter predictions and ground-truth labels. The reported range spans the released settings within each benchmark.
}
\label{tab:instantiations}
\end{table}

\begin{itemize}[leftmargin=1.5em]

\item \cifar: The voter pool consists of ResNet-20~\cite{he2016resnet}, WRN-28-10~\cite{zagoruyko2016wideresnet}, DeiT3-Small~\cite{touvron2022deit3}, and CLIP ViT-B/32~\cite{radford2021clip}, spanning convolutional networks, vision transformers, and multimodal foundation models. Corruption types are drawn from CIFAR-C~\cite{hendrycks2019corruptions}: 3 Noise (\ctype{gaussian}, \ctype{shot}, \ctype{impulse}), 4 Blur (\ctype{defocus}, \ctype{glass}, \ctype{motion}, \ctype{zoom}), 3 Weather (\ctype{fog}, \ctype{frost}, \ctype{snow}), 1 Geometric (\ctype{elastic}), and 4 Digital (\ctype{brightness}, \ctype{contrast}, \ctype{jpeg}, \ctype{pixelate}) corruptions. Combining 15 corruption types with three severity levels yields 45 benchmark settings.

\item \mnist: The voter pool consists of LeNet-5~\cite{lecun1998gradient}, ResNet-20~\cite{he2016resnet}, a multilayer perceptron (MLP), and DeiT3-Small~\cite{touvron2022deit3}. Corruption types are drawn from MNIST-C~\cite{mu2021mnistc}: 3 Noise (\ctype{shot}, \ctype{impulse}, \ctype{spatter}), 2 Blur (\ctype{glass}, \ctype{motion}), 4 Geometric (\ctype{rotation}, \ctype{shear}, \ctype{translation}, \ctype{scaling}), 2 Weather/Digital (\ctype{fog}, \ctype{brightness}), and 4 Structural (\ctype{canny-edges}, \ctype{dotted-line}, \ctype{stripe}, \ctype{zigzag}) corruptions. The Structural corruptions are binary operators and therefore contribute one setting each rather than three severity levels. Consequently, the 11 continuous-severity corruption types contribute 33 settings, while the 4 Structural corruptions contribute 4 additional settings, yielding 37 candidate settings in total. In practice, several low-impact corruption settings produce negligible label uncertainty and are excluded from the final benchmark suite, resulting in 30 released settings.

\item \adult: The voter pool consists of XGBoost~\cite{chen2016xgboost}, CatBoost~\cite{prokhorenkova2018catboost}, an RTDL-tuned MLP~\cite{gorishniy2021revisiting}, FT-Transformer~\cite{gorishniy2021revisiting}, and TabPFN~\cite{hollmann2025tabpfn}. Corruption types include 3 Missing (\ctype{mcar}, \ctype{mar}, \ctype{mnar}) and 2 Value (\ctype{gaussian-noise}, \ctype{scaling}) corruptions. Combining five corruption types with three severity levels yields 15 benchmark settings. For \ctype{mar}, corruption probabilities depend on the observed \texttt{sex} attribute, while for \ctype{mnar} they depend on the values being corrupted. Severity controls the fraction of affected rows. For value corruptions, \ctype{gaussian-noise} adds zero-mean Gaussian noise with $\sigma\in{2.0,3.5,5.0}$ and \ctype{scaling} multiplies numerical values by factors of ${10,100,1000}$ for severity levels $\ell\in{1,3,5}$.
\end{itemize}

For each corruption setting $\tau$, the algorithm produces a benchmark instance $B_\tau$ as defined in Section~\ref{sec:framework}. Consequently, each benchmark suite consists of benchmark instances generated under different ambiguity mechanisms and severity levels. Severity is implemented in a corruption-specific manner, with higher levels corresponding to stronger degradation. For image datasets, we adopt the severity parameterizations provided by CIFAR-C~\cite{hendrycks2019corruptions} and MNIST-C~\cite{mu2021mnistc}; for tabular datasets, we follow the corruption operators introduced by Jenga~\cite{jenga2023}. Complete corruption parameters, benchmark construction code, and reproducibility artifacts are available at \url{https://github.com/sh-islam/ciln-bench}.
 
\subsection{Label Distributions and Confusion Structure}
\label{sec:label_dist}

\begin{figure}[t]
\centering

\begin{subfigure}[t]{0.48\columnwidth}
\centering
\includegraphics[width=\linewidth]{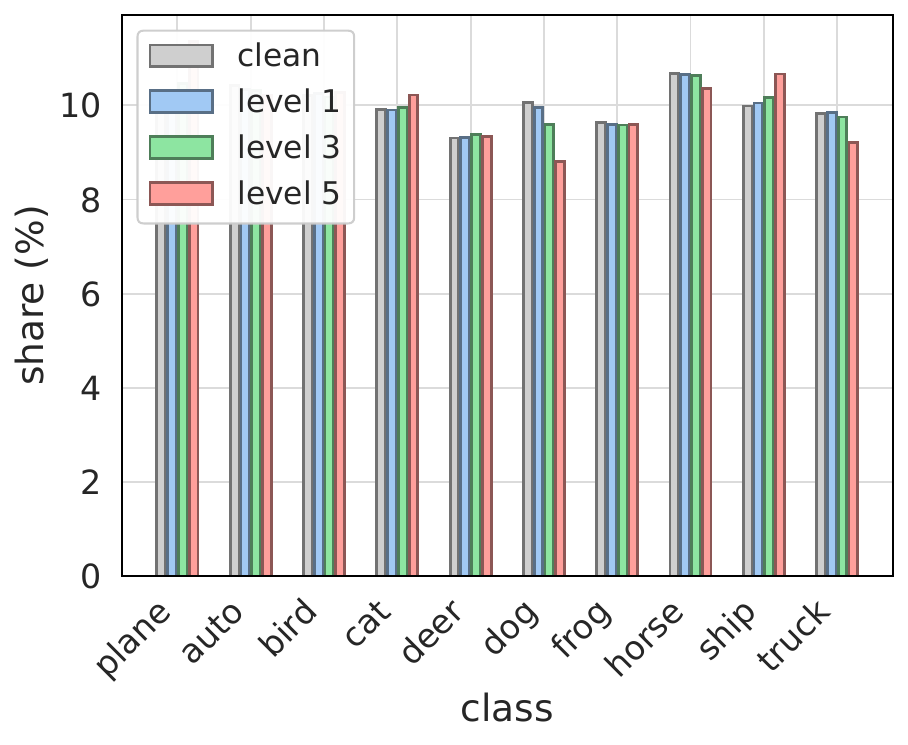}
\caption{\cifar{} \ctype{brightness}}
\label{fig:dist_cifar_brightness}
\end{subfigure}\hfill
\begin{subfigure}[t]{0.48\columnwidth}
\centering
\includegraphics[width=\linewidth]{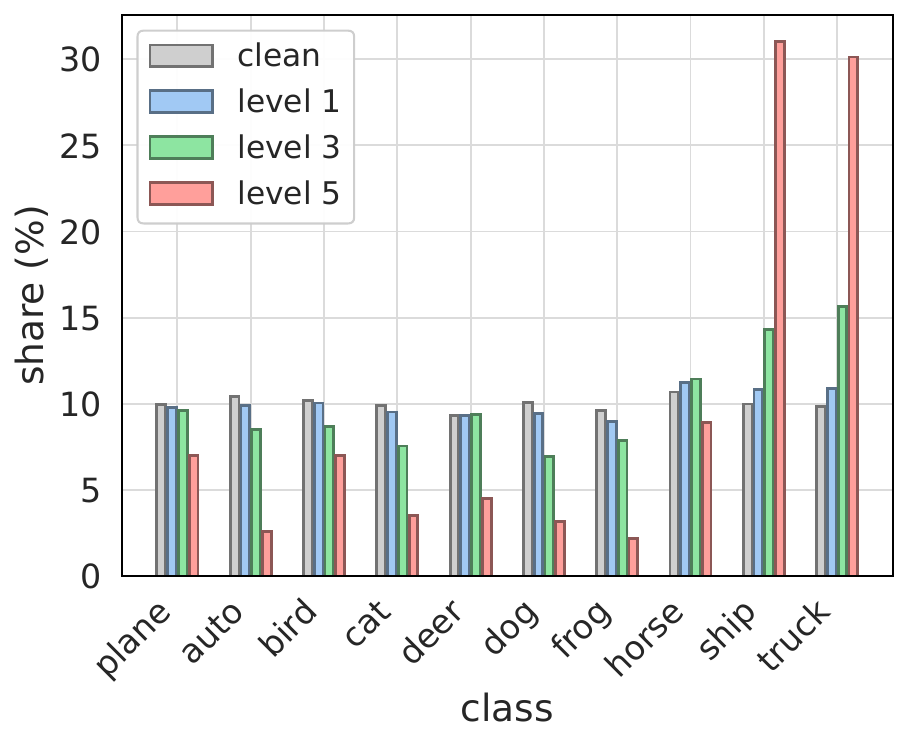}
\caption{\cifar{} \ctype{pixelate}}
\label{fig:dist_cifar_pixelate}
\end{subfigure}

\begin{subfigure}[t]{0.48\columnwidth}
\centering
\includegraphics[width=\linewidth]{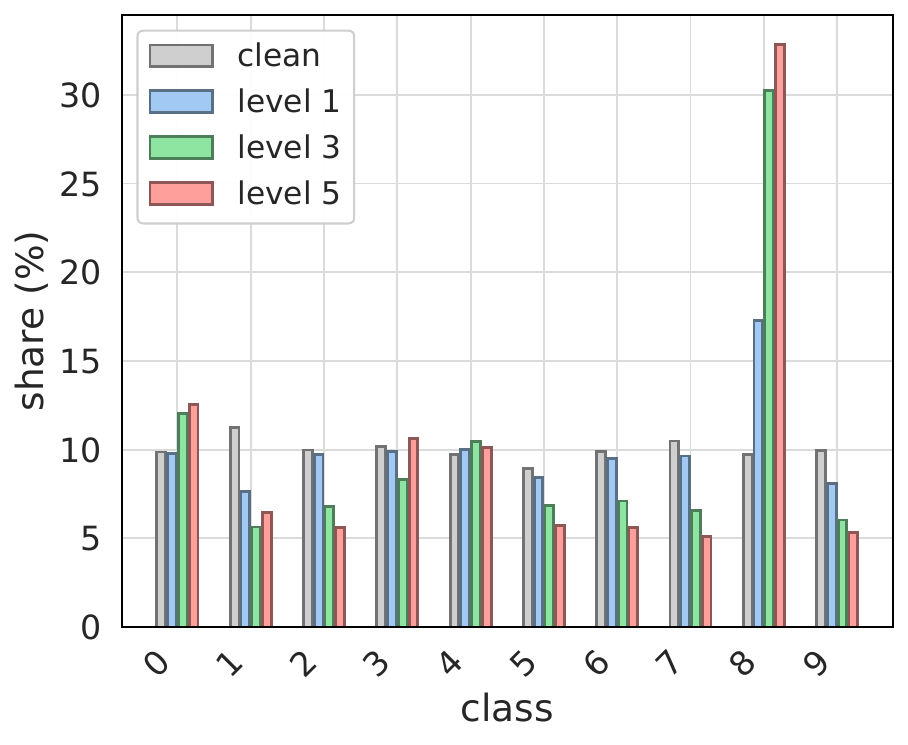}
\caption{\mnist{} \ctype{brightness}}
\label{fig:dist_mnist_brightness}
\end{subfigure}\hfill
\begin{subfigure}[t]{0.48\columnwidth}
\centering
\includegraphics[width=\linewidth]{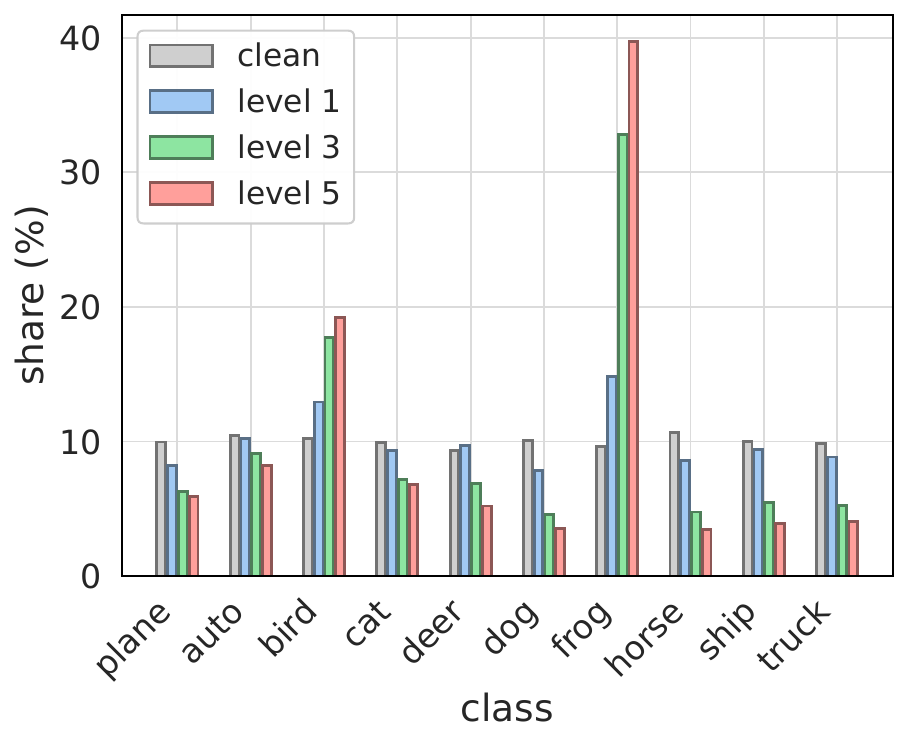}
\caption{\cifar{} \ctype{gaussian}}
\label{fig:dist_cifar_gaussian}
\end{subfigure}

\caption{Examples of corruption-induced label distributions for image datasets across severity levels.}
\label{fig:severity_distributions}
\end{figure}

\begin{figure}[t]
\centering

\begin{subfigure}[t]{0.48\columnwidth}
\centering
\includegraphics[width=\linewidth]{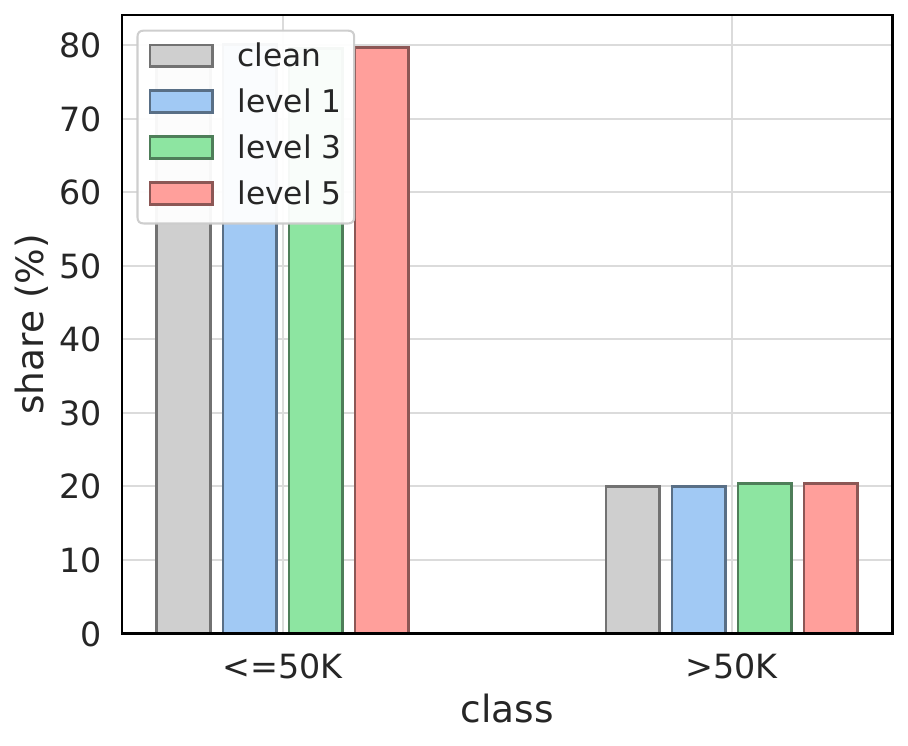}
\caption{\adult{} \ctype{mcar}}
\label{fig:dist_adult_mcar}
\end{subfigure}\hfill
\begin{subfigure}[t]{0.48\columnwidth}
\centering
\includegraphics[width=\linewidth]{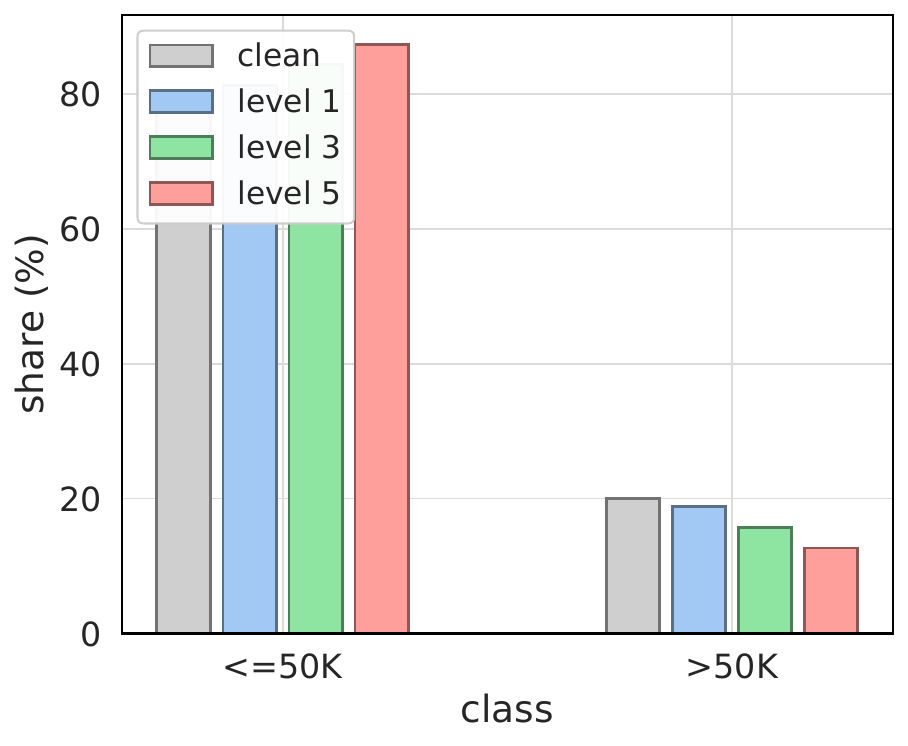}
\caption{\adult{} \ctype{mnar}}
\label{fig:dist_adult_mnar}
\end{subfigure}

\caption{Examples of corruption-induced label distributions for \adult{} under different missing-data mechanisms.}
\label{fig:adult_missingness}
\end{figure}

\begin{figure}[t]
\centering
\begin{subfigure}[t]{0.48\columnwidth}
\centering
\includegraphics[width=\linewidth]{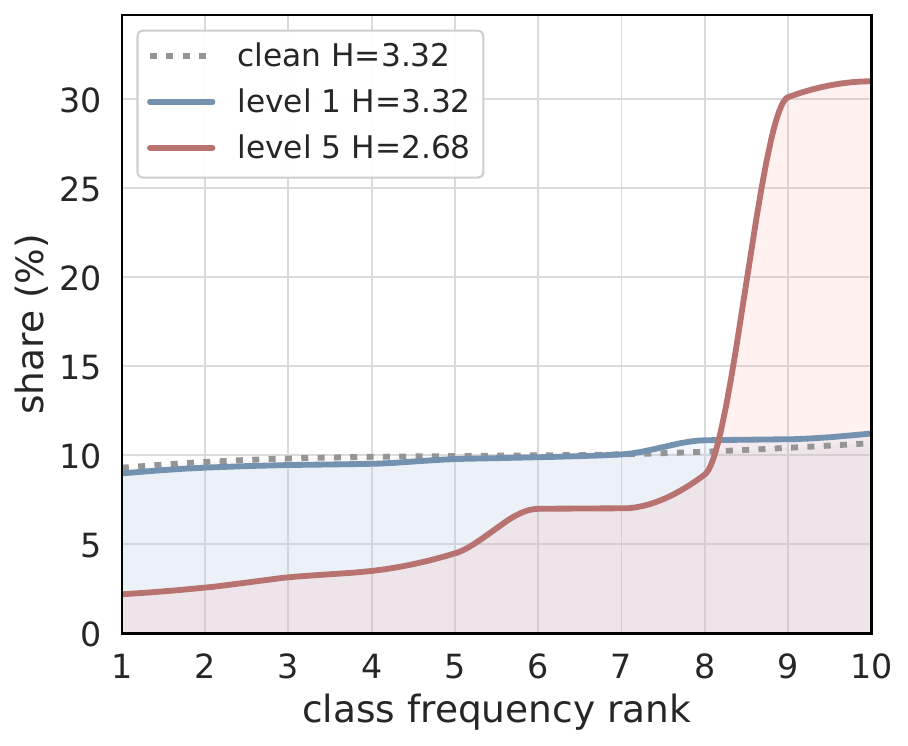}
\caption{\cifar{} \ctype{pixelate}}
\end{subfigure}\hfill
\begin{subfigure}[t]{0.48\columnwidth}
\centering
\includegraphics[width=\linewidth]{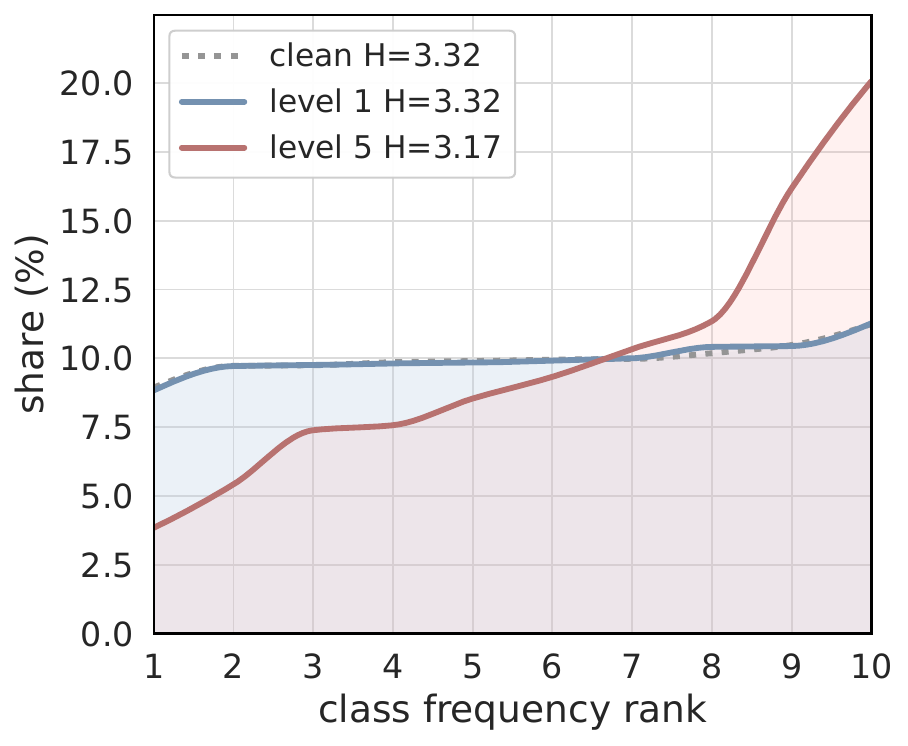}
\caption{\mnist{} \ctype{rotation}}
\end{subfigure}

\begin{subfigure}[t]{0.48\columnwidth}
\centering
\includegraphics[width=\linewidth]{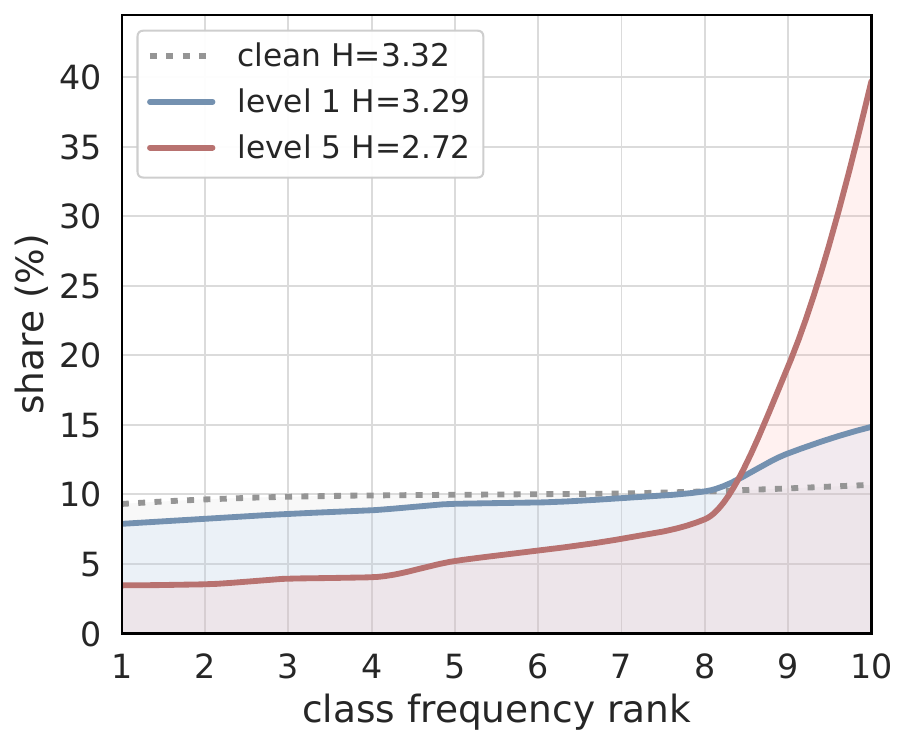}
\caption{\cifar{} \ctype{gaussian}}
\end{subfigure}\hfill
\begin{subfigure}[t]{0.48\columnwidth}
\centering
\includegraphics[width=\linewidth]{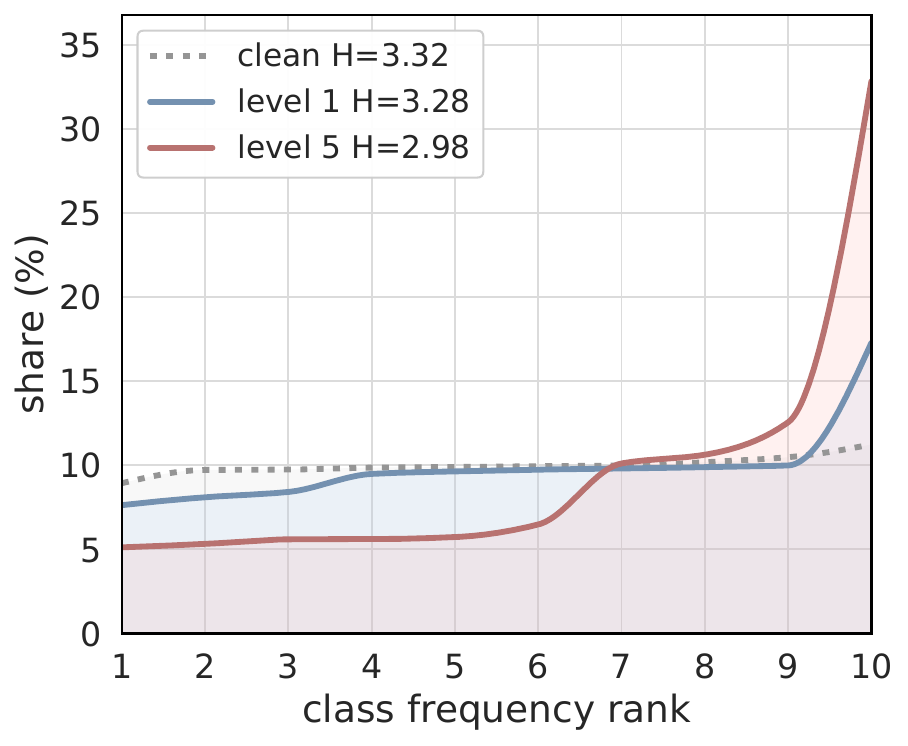}
\caption{\mnist{} \ctype{brightness}}
\end{subfigure}

\caption{Sorted class-frequency profiles for selected corruption settings. Each panel compares clean labels with low- and high-severity corruptions, with label entropy reported in the legend. Lower entropy indicates stronger concentration and greater induced imbalance.}
\label{fig:sorted_frequency_profiles}
\end{figure}

We next examine how controlled input corruptions change the resulting label distributions. This analysis characterizes the type of benchmark instances produced by \cilnbench{}: corruptions may preserve the original class balance, introduce moderate distributional shifts, or concentrate noisy labels around a small number of \emph{attractor classes}. Figures~\ref{fig:severity_distributions} and \ref{fig:adult_missingness} show representative examples, and Figure~\ref{fig:sorted_frequency_profiles} summarizes the induced imbalance through sorted class-frequency profiles and label entropy.

The main observation is that corruption-induced label noise in \cilnbench{} is structured rather than arbitrary. Corruption type determines the \emph{direction} of the shift: some corruptions leave the label distribution close to the clean baseline, while others pull probability mass toward specific classes. For example, Figure~\ref{fig:severity_distributions}\subref{fig:dist_cifar_brightness} shows that \ctype{brightness} has little effect on the \cifar{} label distribution, whereas Figure~\ref{fig:severity_distributions}\subref{fig:dist_cifar_pixelate} shows that \ctype{pixelate} creates a strong shift toward a small set of classes. Severity then controls the \emph{magnitude} of this effect. As severity increases, distributions often become more concentrated, entropy decreases, and class imbalance becomes stronger. Thus, corruption type determines the structure of the induced noise, while severity controls its intensity.

The effect of corruption is also dataset-dependent. The same corruption mechanism can produce substantially different distributional shifts on different datasets. For instance, Figures~\ref{fig:severity_distributions}\subref{fig:dist_cifar_brightness} and \ref{fig:severity_distributions}\subref{fig:dist_mnist_brightness} show that \ctype{brightness} produces little imbalance on \cifar{}, but substantially concentrates \mnist{} labels toward a small subset of classes. Similarly, Figures~\ref{fig:adult_missingness}\subref{fig:dist_adult_mcar} and \ref{fig:adult_missingness}\subref{fig:dist_adult_mnar} show that different missing-data mechanisms can have markedly different effects on class balance in \adult{}. These examples demonstrate that corruption effects are not universal properties of the corruption alone; they depend on how the corruption interacts with the underlying structure of the dataset.

Attractor classes provide an extreme manifestation of this dataset-dependent behavior. Under some corruption--dataset pairs, many different true classes are mapped toward the same predicted class. In \cifar{}, high-severity noise corruptions often increase the share of \emph{frog}, while blur corruptions frequently increase the share of \emph{cat}. In \mnist{}, pixel-degradation corruptions such as \ctype{impulse-noise} and \ctype{brightness} tend to pull labels toward digit 8, whereas geometric corruptions such as \ctype{rotation} tend to pull labels toward digit 2. In \adult{}, missingness often pushes predictions toward the majority class, while \ctype{scaling} can shift predictions in the opposite direction. These attractor patterns reveal that corruption-induced noise changes not only the amount of label noise, but also its class-level geometry.

Figure~\ref{fig:sorted_frequency_profiles} provides a complementary view of these effects through sorted class-frequency profiles. Consistent with the severity trends observed above, settings such as \cifar{} \ctype{pixelate}, \cifar{} \ctype{gaussian-noise}, and \mnist{} \ctype{brightness} become progressively more skewed as severity increases, producing lower-entropy label distributions. In contrast, settings with weaker distributional effects remain closer to the clean profile. This visualization reinforces that severity acts as a control knob for the strength of the induced imbalance, while the corruption--dataset pair determines its specific form.

These results show that \cilnbench{} provides control over both the amount and the structure of label noise. Corruption type determines the direction of the distributional shift, severity determines its strength, and the dataset governs how these effects are realized. This characterization helps explain the experimental results in Section~\ref{sec:exp}: methods that perform well on rate-matched \plidn{} can fail on \cilnbench{} because the challenge is not merely higher noise rates, but fundamentally different noise directions and imbalance structures.

\section{Experimental Evaluation}
\label{sec:exp}

Our evaluation has three objectives. First, we investigate whether controlled corruptions produce genuine instance-dependent label noise and whether the degree of instance dependence can be systematically influenced through the choice of corruption type and severity. Second, we examine how closely the resulting label distributions resemble human annotation uncertainty, assessing whether corruption-induced ambiguity can serve as a realistic proxy for human disagreement. Third, we evaluate the usefulness of the benchmark as a stress test for noisy-label learning by studying how state-of-the-art methods behave under corruption-induced label noise. Together, these experiments assess the realism, controllability, and practical value of the proposed benchmark.

\subsection{Experimental Setup}

We first describe the competing benchmark used for comparison and then introduce the evaluation measures used throughout the experiments.

\subsubsection{Competing Benchmark Generation Methods}
\label{sec:compete}

We compare \cilnbench{} against \plidn~\cite{gu2022idn}, a widely used benchmark for instance-dependent label noise. \plidn{} is available only for \cifar{} and therefore serves as our primary reference benchmark for evaluating instance dependence, similarity to human annotation uncertainty, and downstream learning behavior. No comparable benchmark release exists for \mnist{} or \adult{}, so for those datasets we report the absolute characteristics of the generated benchmarks.

The two benchmarks differ in how label noise is generated. \plidn{} produces noisy labels by using imperfect model raters on clean images, whereas \cilnbench{} generates label uncertainty by applying controlled corruptions to the input and aggregating predictions from a voter pool. Consequently, \plidn{} controls noise through the properties of the raters, while \cilnbench{} controls noise through the type and severity of corruption.

\plidn{} is released in three versions corresponding to low-, medium-, and high-noise regimes, which we denote by \plidnL{}, \plidnM{}, and \plidnH{}, respectively. Unless otherwise stated, \plidn{} refers to the benchmark generation method without reference to a particular noise level. In contrast, \cilnbench{} provides a broader collection of benchmark settings spanning multiple corruption mechanisms and severity levels, allowing users to generate benchmarks with varying degrees and structures of label noise. When comparing \cilnbench{} against \plidn{}, we compare each \plidn{} release with the \cilnbench{} settings having the closest observed noise rates. For reference, \plidnL{}, \plidnM{}, and \plidnH{} correspond to approximately 11\%, 19\%, and 48\% label noise, respectively.

In addition to the default benchmark release, \cilnbench{} supports an optional \emph{clean-start} configuration (Algorithm~\ref{alg:cilnbench}), which retains only instances that are classified correctly by all voters before corruption. We denote the default benchmark by \cilnbenchS{} and the clean-start variant by \cilnbenchC{}. Unless stated otherwise, \cilnbench{} refers to \cilnbenchS{}. The clean-start variant isolates corruption-induced uncertainty from pre-existing ambiguity in the original data. Unless otherwise stated, results are reported on \cilnbenchS{}, and selected experiments are repeated on \cilnbenchC{} to assess the robustness of the findings. Because \cilnbenchC{} removes instances that already exhibit voter disagreement before corruption, it typically produces lower noise rates than the corresponding \cilnbenchS{} setting.

\subsubsection{Evaluation Measures and Evaluation Scenarios}
\label{sec:measures}

We evaluate the generated benchmarks from three perspectives: (i) the degree of instance dependence in the induced label noise, (ii) similarity to human annotation uncertainty, and (iii) downstream learning performance.

To quantify instance dependence, we introduce \emph{Noise Transition Heterogeneity} (\nth), which measures the variability of instance-specific noise transition distributions within each class. Under class-conditional noise, all instances belonging to the same class share the same transition distribution, resulting in $\nth=0$. Larger values indicate stronger instance-dependent variation.

Let $\bar{p}_i(\tilde{y}\mid y=y_i)$ denote the aggregated transition distribution of instance $i$, and let
\[
\bar{p}(\tilde{y}\mid y=k)
=
\frac{1}{|I_k|}
\sum_{i\in I_k}
\bar{p}_i(\tilde{y}\mid y=k)
\]
denote the average transition distribution for class $k$, where $I_k=\{i \mid y_i=k\}$. We define \nth{} as
\[
\nth(B_\tau)
=
\frac{1}{N}
\sum_{k=1}^{K}
\sum_{i\in I_k}
\left\|
\bar{p}_i(\tilde{y}\mid y=k)
-
\bar{p}(\tilde{y}\mid y=k)
\right\|_2^2 .
\]

Intuitively, \nth{} measures how much the transition behavior of individual instances deviates from the average transition behavior of their class. A value of zero indicates class-conditional noise, while larger values indicate stronger instance-dependent label noise.

To evaluate similarity to human annotation uncertainty, we compare the generated label distributions against CIFAR-10H using total variation (TV) distance. Lower TV values indicate greater similarity to human label distributions.

For downstream evaluation, we train classifiers using standard empirical risk minimization (ERM), Co-Teaching~\cite{han2018coteaching}, and DivideMix~\cite{li2020dividemix}, and report clean-test accuracy. Section~\ref{sec:rw_noisy_label_learning} provides a brief overview of Co-Teaching and DivideMix. To ensure a fair comparison with \plidn{}, we follow the same protocol for both benchmarks: a single voter (or rater) is sampled uniformly at random for each instance, and its predicted label is used as the training label. We consider two training scenarios. In \emph{clean-img}, learners are trained on clean images paired with corruption-induced labels. In \emph{noisy-img}, learners are trained on the corrupted images paired with the same labels. Comparing the two scenarios isolates the effect of label noise from the additional effect of corrupted inputs.

\subsection{Experimental Results}

We present the results according to the three evaluation objectives: instance dependence, similarity to human annotation uncertainty, and downstream learning performance.

\begin{figure}[t]
\centering
\begin{subfigure}[t]{0.48\columnwidth}
  \centering
  \includegraphics[width=\linewidth]{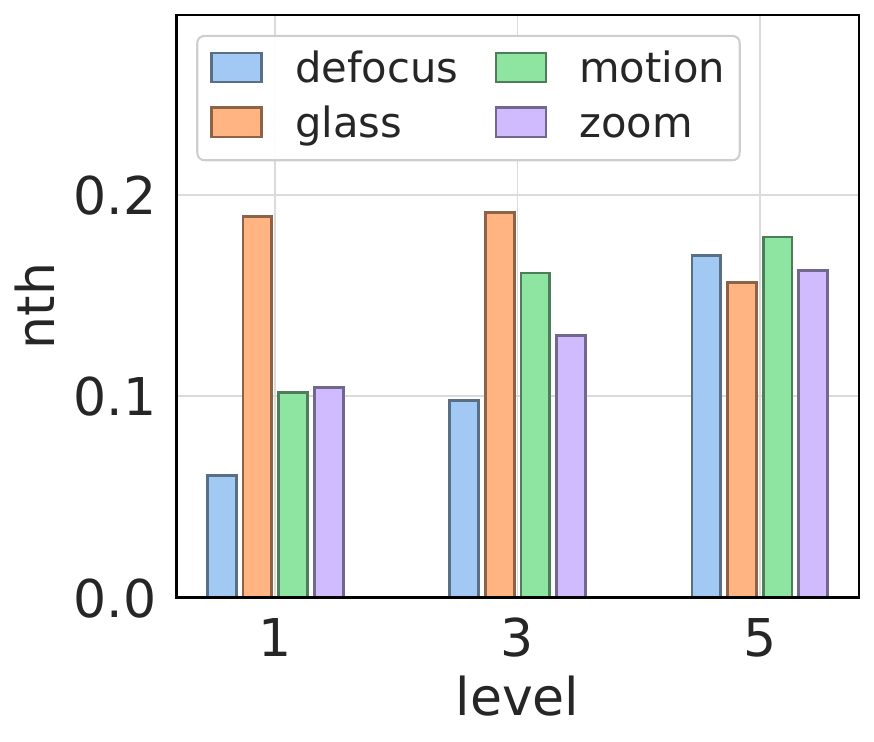}
  \caption{\cfamily{Blur}}
\end{subfigure}\hfill
\begin{subfigure}[t]{0.48\columnwidth}
  \centering
  \includegraphics[width=\linewidth]{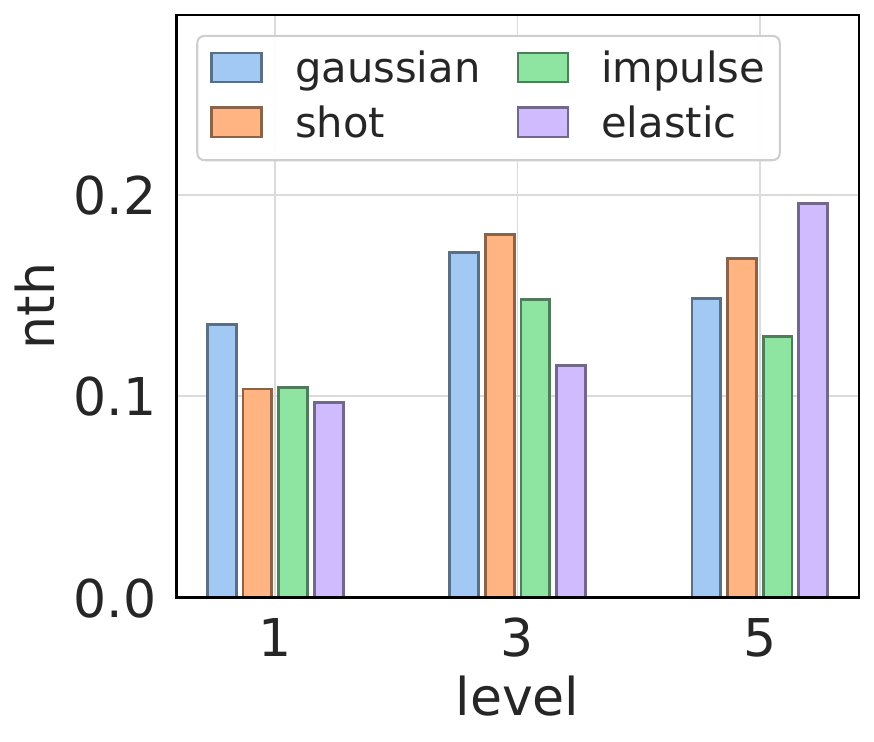}
  \caption{\cfamily{Noise} and \cfamily{Geometric}}
\end{subfigure}

\begin{subfigure}[t]{0.48\columnwidth}
  \centering
  \includegraphics[width=\linewidth]{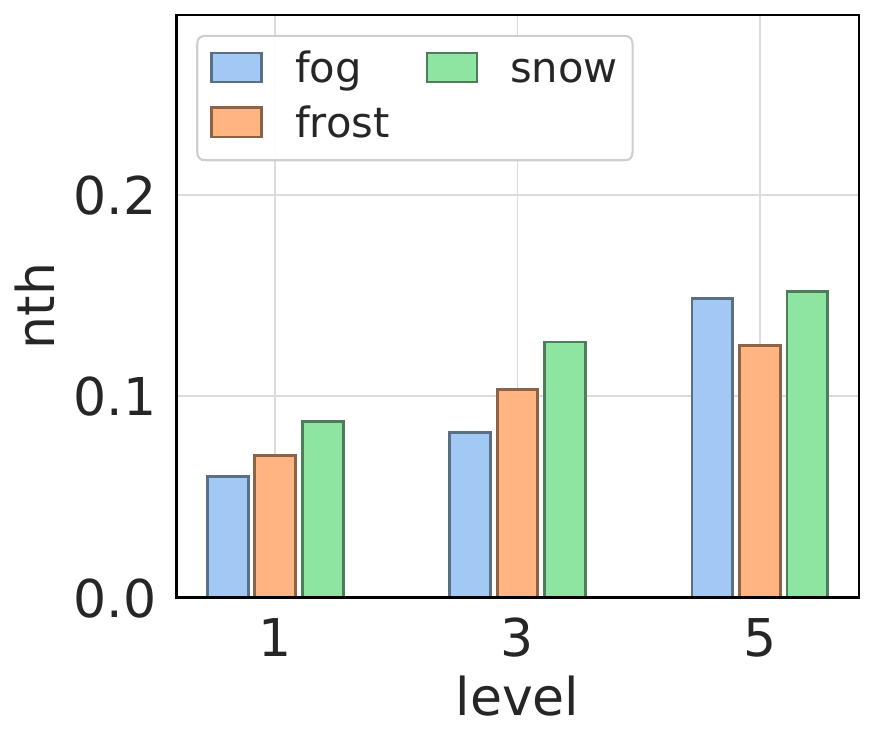}
  \caption{\cfamily{Weather}}
\end{subfigure}\hfill
\begin{subfigure}[t]{0.48\columnwidth}
  \centering
  \includegraphics[width=\linewidth]{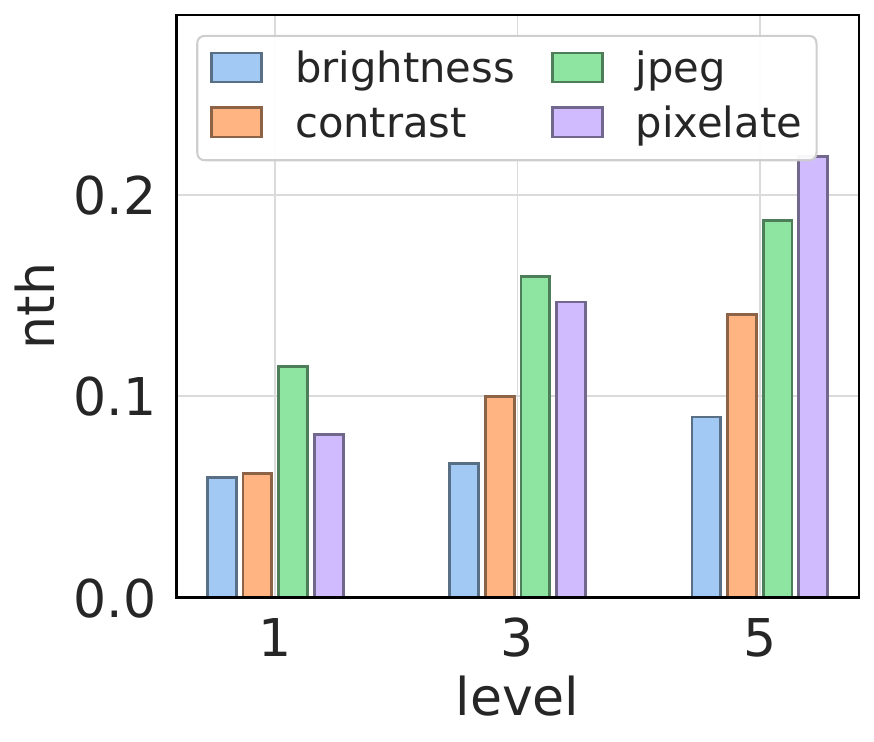}
  \caption{\cfamily{Digital}}
\end{subfigure}
\caption{CIFAR-10 nth across corruption super-families and severities. Each panel groups corruption sub-families within a super-family; bars are shown for levels 1, 3, and 5.}
\label{fig:cifar_severity_bars}
\end{figure}

\subsubsection{Strength of Instance Dependence}
\label{sec:exp1}

We evaluate instance dependence using \nth. Recall that \nth{} measures the variability of instance-specific transition distributions within each class. A value of zero corresponds to class-conditional noise, while larger values indicate stronger instance dependence.

Figure~\ref{fig:cifar_severity_bars} summarizes \nth{} across the 45 \cifar{} corruption settings. All settings produce positive \nth{} values, indicating that the induced label noise is instance-dependent rather than class-conditional. The amount of instance dependence varies substantially across corruption families and severity levels, showing that \cilnbench{} provides explicit control over the strength of the induced IDN. For many corruption families, \nth{} increases with severity at low and moderate levels, but eventually saturates or decreases at high severity. Once a corruption becomes sufficiently strong, many examples exhibit similar prediction patterns, reducing variability across instances. As a result, higher noise rates do not necessarily imply stronger instance dependence.

Table~\ref{tab:cifar_overview} compares \cilnbench{} and \plidn{} at comparable noise rates. At low noise rates, \plidn{} generally exhibits stronger instance dependence. At medium noise rates, the two benchmarks are comparable, with some \cilnbench{} settings exceeding \plidn{}. At high noise rates, representative \cilnbench{} settings match or exceed the \nth{} values observed in \plidn{}. Beyond matching \plidn{}, \cilnbench{} provides a direct mechanism for controlling instance dependence through corruption type and severity. In contrast, \plidn{} increases noise by progressively degrading the quality of its rater pool. Consequently, the instance dependence observed in \cilnbench{} is driven by controlled input ambiguity, whereas the instance dependence observed in \plidn{} may also reflect variation in rater quality.

No \plidn{} reference benchmark exists for \mnist{} or \adult{}, so we report the absolute behavior of \cilnbench{}. For both datasets, all evaluated settings produce positive \nth{} values, indicating consistent instance-dependent behavior. On \mnist{}, the strongest instance dependence is observed for \cfamily{Geometric}, such as \rotate{} and \shear{}, whose effects depend heavily on the shape of the underlying digit. On \adult{}, the strongest instance dependence is observed for the \scaling{} corruption family. These results demonstrate that \cilnbench{} extends beyond image data and remains effective across different data modalities.

Table~\ref{tab:cifar_overview} compares \nth{} under the standard benchmark and the clean-start variant. As expected, clean-start generally produces smaller \nth{} values because it removes examples that already exhibit uncertainty before corruption and therefore reduces the effective noise rate. However, the qualitative conclusions remain unchanged. In particular, \nth{} remains positive across all evaluated settings, and the strongest instance-dependent settings remain the strongest under both benchmark variants. This indicates that the observed instance dependence is driven by the corruption process rather than by pre-existing ambiguity in the underlying data.

\begin{table*}[t]
\centering
\small
% \begin{adjustbox}{max width=\textwidth}
\begin{tabular}{@{}l l rrr r r r r r r@{}}
\toprule
          &            &            &      &      &
\multicolumn{6}{c}{\textbf{Accuracy}} \\
\cmidrule(lr){6-11}
\textbf{Benchmark} & \textbf{Corr. ($\ell$)} & \textbf{Noise Rate} & \textbf{\nth} & \textbf{TV} &
\textbf{ERM-C} & \textbf{ERM-N} &
\textbf{CoT-C} & \textbf{CoT-N} &
\textbf{DM-C} & \textbf{DM-N} \\
\midrule

% ---- LOW TIER ($\sim 11\%$ noise) ----
\plidnL                        & --        & $10.8\%$ & $0.088$ & $0.180\,\pm\,0.005$ & $84.4$ & -- & $85.6$ & -- & $69.2$ & -- \\
\midrule
\multirow{2}{*}{\cilnbenchS{}}
& \ctype{frost} (1)
& $10.1\%$ & $0.070$ & $0.133\,\pm\,0.002$
& $85.0$ & $84.6$ & $86.8$ & $85.9$ & $71.9$ & $69.0$ \\
& \ctype{pixelate} (1)
& $11.7\%$ & $0.081$ & $0.146\,\pm\,0.002$
& $84.0$ & $84.2$ & $86.4$ & $85.9$ & $70.6$ & $69.0$ \\
\midrule
\cilnbenchC{} & \ctype{frost} (1)    & $2.8\%$ & $0.016$ & $0.051\,\pm\,0.002$ & -- & -- & -- & -- & -- & -- \\
              & \ctype{pixelate} (1) & $4.1\%$ & $0.025$ & $0.061\,\pm\,0.001$ & -- & -- & -- & -- & -- & -- \\
\specialrule{1pt}{2pt}{3pt}

% ---- MED TIER ($\sim 19\%$ noise) ----
\plidnM{} & -- & $19.4\%$ & $0.121$ & $0.301\,\pm\,0.008$
& $79.1$ & -- & $82.3$ & -- & $68.8$ & --\\
\midrule

\multirow{3}{*}{\cilnbenchS{}}
& \ctype{elastic} (3)
& $17.4\%$ & $0.115$ & $0.209\,\pm\,0.002$
& $80.4$ & $80.0$ & $83.9$ & $82.5$ & $68.2$ & $66.8$ \\
& \ctype{snow} (3)
& $19.4\%$ & $0.127$ & $0.227\,\pm\,0.002$
& $78.9$ & $80.8$ & $84.3$ & $83.1$ & $67.6$ & $65.5$ \\
& \ctype{gaussian} (1)
& $21.5\%$ & $0.135$ & $0.247\,\pm\,0.002$
& $78.2$ & $78.4$ & $82.4$ & $81.8$ & $65.7$ & $67.6$ \\
\midrule

\multirow{3}{*}{\cilnbenchC{}}
& \ctype{elastic} (3)
& $10.1\%$ & $0.059$ & $0.128\,\pm\,0.002$
& -- & -- & -- & -- & -- & -- \\
& \ctype{snow} (3)
& $12.5\%$ & $0.078$ & $0.152\,\pm\,0.002$
& -- & -- & -- & -- & -- & -- \\
& \ctype{gaussian} (1)
& $14.4\%$ & $0.088$ & $0.175\,\pm\,0.002$
& -- & -- & -- & -- & -- & -- \\
\specialrule{1pt}{2pt}{3pt}

% ---- HIGH TIER ($\sim 48\%$ noise) ----
\plidnH{} & -- & $47.8\%$ & $0.152$ & $0.742\,\pm\,0.019$
& $63.3$ & -- & $65.8$ & -- & $57.8$ & -- \\
\midrule

\multirow{5}{*}{\cilnbenchS{}}
& \ctype{contrast} (5)
& $47.3\%$ & $0.140$ & $0.486\,\pm\,0.002$
& $59.1$ & $51.4$ & $42.8$ & $22.1$ & $45.2$ & $22.3$ \\
& \ctype{gaussian} (3)
& $49.2\%$ & $0.171$ & $0.507\,\pm\,0.002$
& $60.6$ & $61.3$ & $44.9$ & $44.7$ & $45.8$ & $47.0$ \\
& \ctype{shot} (5)
& $54.5\%$ & $0.168$ & $0.555\,\pm\,0.002$
& -- & -- & $28.7$ & $28.6$ & -- & -- \\
& \ctype{pixelate} (5)
& $58.2\%$ & $0.219$ & $0.598\,\pm\,0.002$
& $46.5$ & $49.5$ & $27.1$ & $26.6$ & $37.0$ & $36.4$ \\
& \ctype{glass-blur} (3)
& $64.6\%$ & $0.191$ & $0.653\,\pm\,0.002$
& $41.2$ & $47.8$ & $21.4$ & $20.5$ & $29.9$ & $32.1$ \\
\midrule

\multirow{5}{*}{\cilnbenchC{}}
& \ctype{contrast} (5)
& $43.7\%$ & $0.125$ & $0.457\,\pm\,0.002$
& $57.2$ & $49.3$ & $47.3$ & $24.1$ & -- & -- \\
& \ctype{gaussian} (3)
& $44.7\%$ & $0.157$ & $0.469\,\pm\,0.002$
& $58.7$ & $59.2$ & $46.4$ & $48.7$ & -- & -- \\
& \ctype{shot} (5)
& $51.0\%$ & $0.156$ & $0.524\,\pm\,0.002$
& $52.3$ & $54.2$ & $38.4$ & $36.2$ & -- & -- \\
& \ctype{pixelate} (5)
& $52.7\%$ & $0.200$ & $0.554\,\pm\,0.002$
& -- & -- & $28.6$ & $28.8$ & -- & -- \\
& \ctype{glass-blur} (3)
& $61.8\%$ & $0.186$ & $0.632\,\pm\,0.003$
& -- & -- & $22.3$ & $22.1$ & -- & -- \\

\bottomrule
\end{tabular}
% \end{adjustbox}
\caption{Comparison of \plidn{} and representative \cilnbench{} settings on \cifar{} at comparable noise rates. Results are shown for both \cilnbenchS{} and \cilnbenchC{}. Empty cells reflect deprioritized runs: under \cilnbenchS{}, ERM and DivideMix were not re-run for \ctype{shot} (5) since the \cfamily{Noise} family was already covered by \ctype{gaussian} (3); under \cilnbenchC{}, ERM and DivideMix were only run on the three closest noise-rate matches to \plidn{}'s high tier, and DivideMix was not re-run under \cilnbenchC{} since Co-Teaching already confirmed the same small-loss failure.}

\label{tab:cifar_overview}
\end{table*}

\begin{figure}[t]
    \centering

    \begin{subfigure}[t]{0.48\linewidth}
        \centering
        \includegraphics[width=\linewidth]{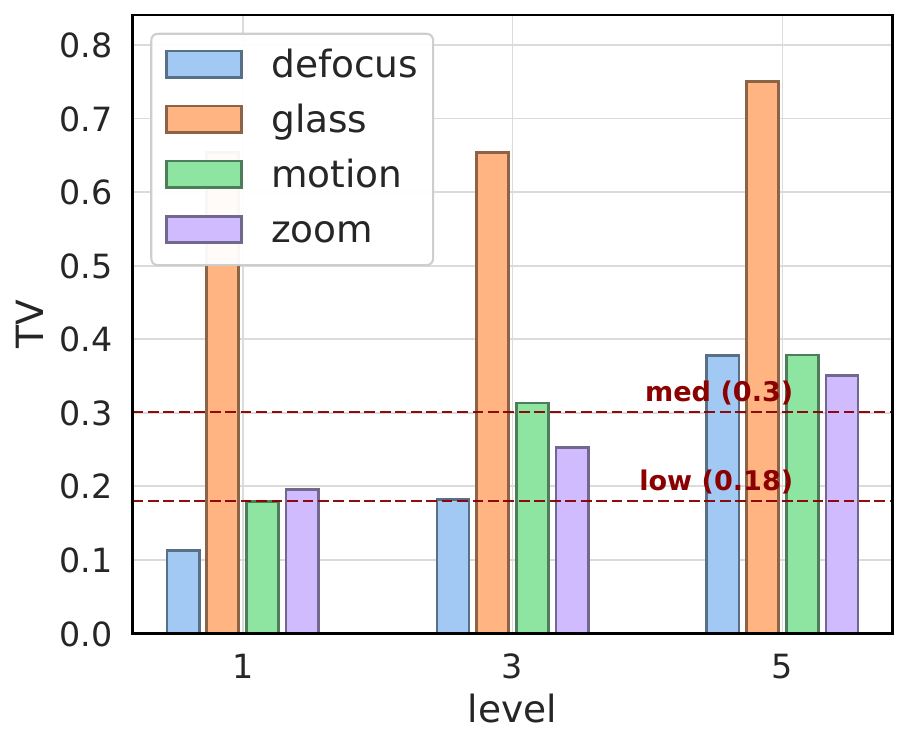}
        \caption{\cfamily{Blur}}
        \label{fig:tv-blur}
    \end{subfigure}
    \hfill
    \begin{subfigure}[t]{0.48\linewidth}
        \centering
        \includegraphics[width=\linewidth]{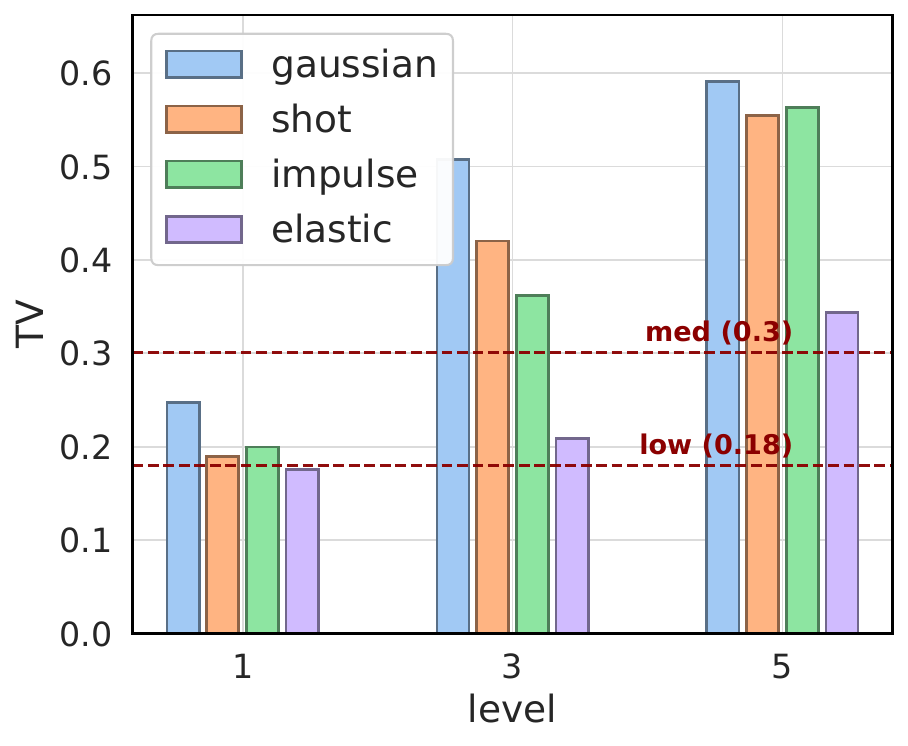}
        \caption{\cfamily{Noise} and \cfamily{Geometric}}
        \label{fig:tv-noise-geometric}
    \end{subfigure}

    \vspace{0.8em}

    \begin{subfigure}[t]{0.48\linewidth}
        \centering
        \includegraphics[width=\linewidth]{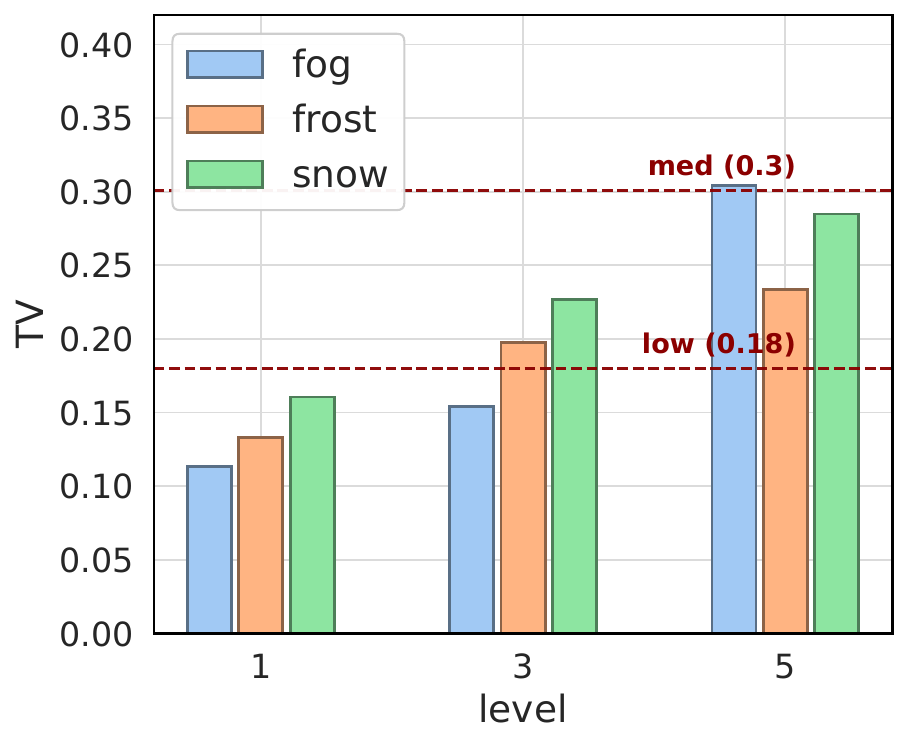}
        \caption{\cfamily{Weather}}
        \label{fig:tv-weather}
    \end{subfigure}
    \hfill
    \begin{subfigure}[t]{0.48\linewidth}
        \centering
        \includegraphics[width=\linewidth]{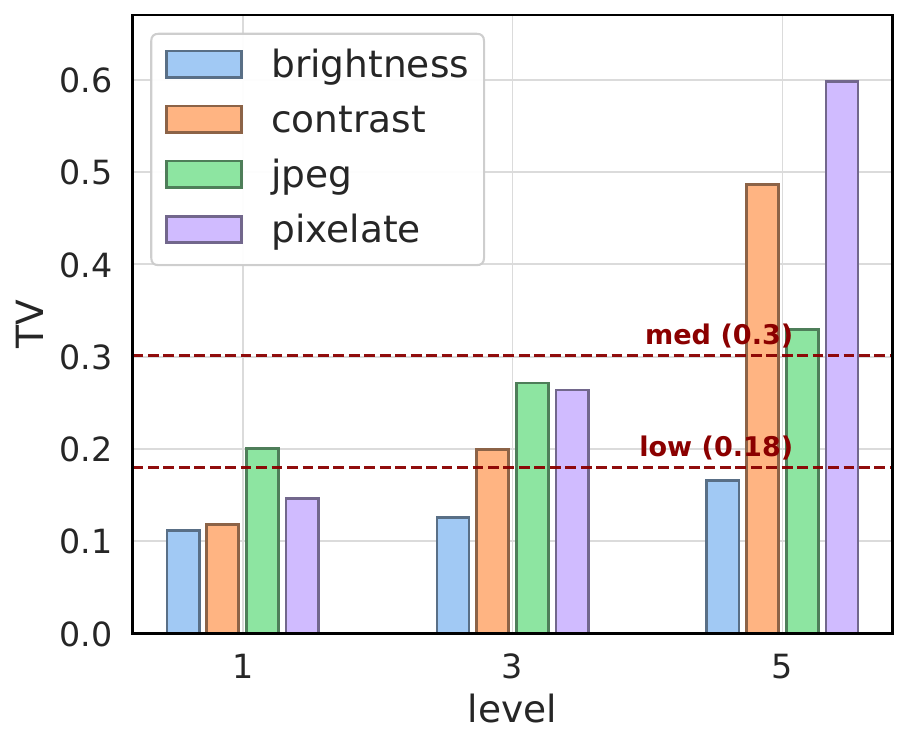}
        \caption{\cfamily{Digital}}
        \label{fig:tv-digital}
    \end{subfigure}

    \caption{Total variation distance to \cifarh{} across corruption families and severity levels. Lower values indicate greater similarity to human annotation uncertainty.}
    \label{fig:tv-cifar10h}
\end{figure}

\begin{figure}[t]
\centering
\includegraphics[width=0.85\columnwidth]{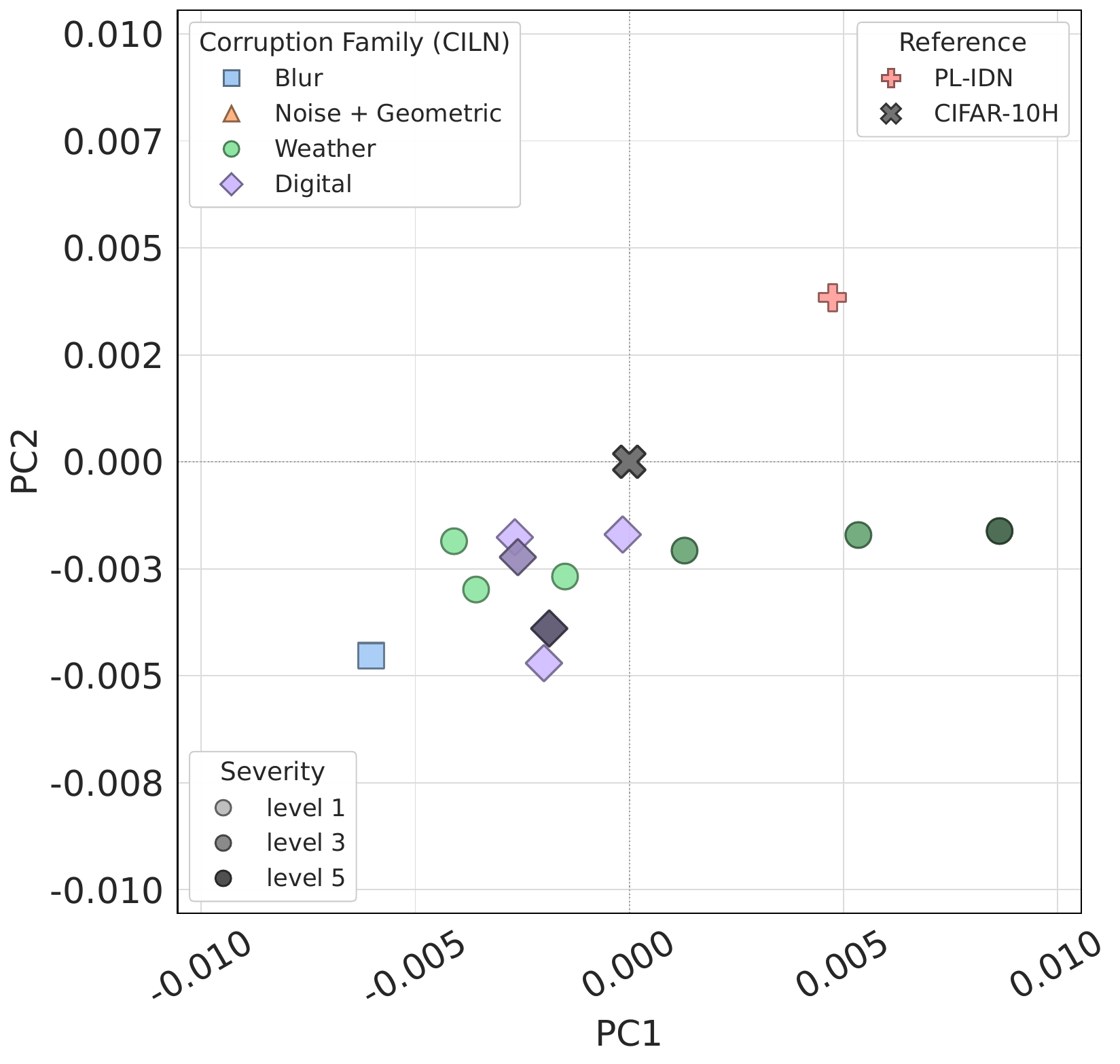}
\vspace{-2mm}
\caption{The 10 \cilnbench{} settings closest to \cifarh{} in PCA space. The reference \cifarh{} distribution is shown as a black times sign and the closets \plidn{} release as a red plus.}
\label{fig:tv_pca_top10}
\end{figure}

\subsubsection{Similarity to Human Annotation Uncertainty}
\label{sec:exp2}

We use \cifarh{} as a reference for human annotation uncertainty and measure the similarity between benchmark-generated label distributions and human label distributions using TV distance. Lower TV values indicate greater similarity to human labeling behavior.

Table~\ref{tab:cifar_overview} shows that \cilnbench{} is consistently closer to \cifarh{} than \plidn{} across all three \plidn{} noise levels. The gap increases as the noise rate grows, indicating that the label distributions produced by \cilnbench{} remain closer to human disagreement patterns even in high-noise settings. 

Figure~\ref{fig:tv-cifar10h} shows how similarity to human labeling varies across corruption families and severity levels. Human-likeness is not uniform across corruption settings. In general, mild corruptions produce label distributions that are closer to human annotation uncertainty, whereas severe corruptions tend to move further away from the \cifarh{} reference. Weather and Digital corruptions consistently achieve the lowest TV distances, while severe Noise and Blur corruptions are less human-like despite still outperforming \plidn{}.

Figure~\ref{fig:tv_pca_top10} provides a complementary view by ranking corruption settings according to their proximity to \cifarh{} in PCA space. The ten closest settings are all low-severity Weather or Digital corruptions, including \bright{}, \frost{}, \snow{}, \fog{}, \contr{}, and \jpeg{}. Notably, none of the three released \plidn{} datasets appears among the ten closest settings. This suggests that mild corruption-induced ambiguity provides a particularly realistic approximation of human labeling uncertainty.

Table~\ref{tab:cifar_overview} also shows that the main conclusion is robust to the clean-start variant. Although clean-start slightly increases TV distance in some low-noise settings by removing naturally ambiguous examples, \cilnbench{} remains consistently closer to \cifarh{} than \plidn{}. At high noise rates, the two benchmark variants produce nearly identical results.

\begin{table}[t]
\centering
\small
\begin{adjustbox}{max width=\columnwidth}
\begin{tabular}{lrrrl}
\toprule
Corr. Type & Rate & Purity & Co-T & Attractor Class \\
\midrule
\plidnH{}                & $47.8\%$ & $38.8\%$ & $65.8$       & truck (+8 pp) \\
\midrule
\ctype{gaussian} (3) & $49.2\%$ & $24.1\%$ & $44.9/44.7$ & frog (+22 pp) \\
\ctype{contrast} (5)       & $47.3\%$ & $26.2\%$ & $42.8/22.1$ & cat (+19 pp) \\
\ctype{pixelate} (5)       & $58.2\%$ & $27.3\%$ & $27.1/26.6$ & truck (+20 pp) \\
\ctype{glass} (3)     & $64.7\%$ & $19.9\%$ & $22.0/21.1$ & frog (+30 pp) \\
\bottomrule
\end{tabular}\end{adjustbox}
\caption{Representative high-noise settings. The attractor class is the class receiving the largest increase in label frequency relative to the clean distribution; the value in parentheses reports the increase in percentage points. Purity is the fraction of examples assigned to the attractor class that truly belong to that class. Co-T reports Co-Teaching clean-test accuracy (clean-img / noisy-img). For comparison, Co-Teaching achieves $65.8\%$ accuracy on \plidnH{}.}
\label{tab:downstream_diagnostics}
\end{table}

\begin{figure}[t]
\centering

\begin{subfigure}[t]{0.32\columnwidth}
  \centering
  \includegraphics[width=\linewidth]{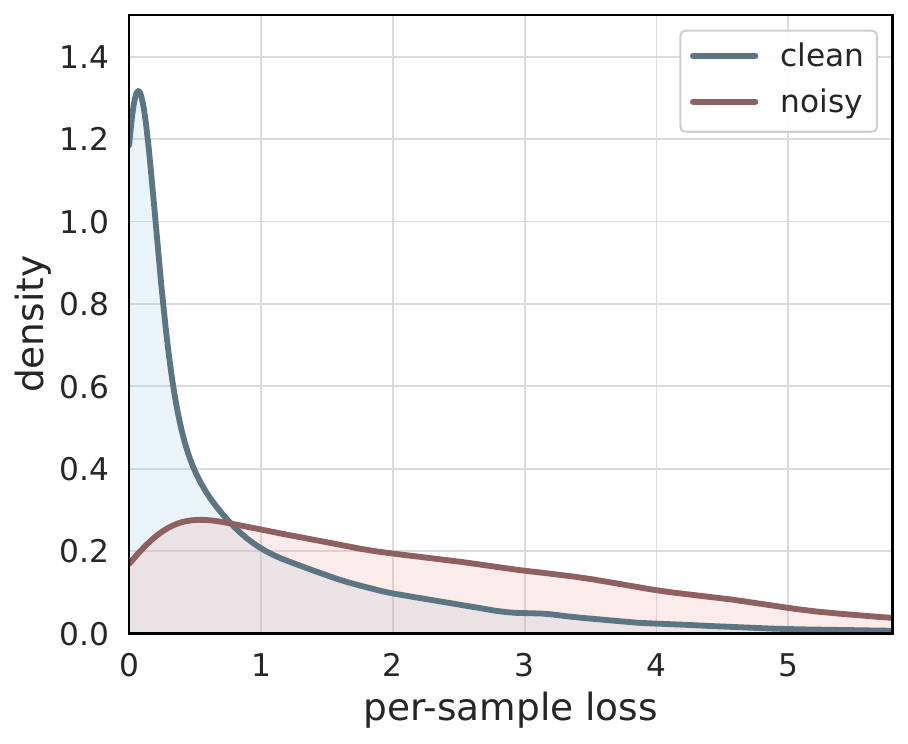}
  \caption{\ctype{frost} (1)}
\end{subfigure}\hfill
\begin{subfigure}[t]{0.32\columnwidth}
  \centering
  \includegraphics[width=\linewidth]{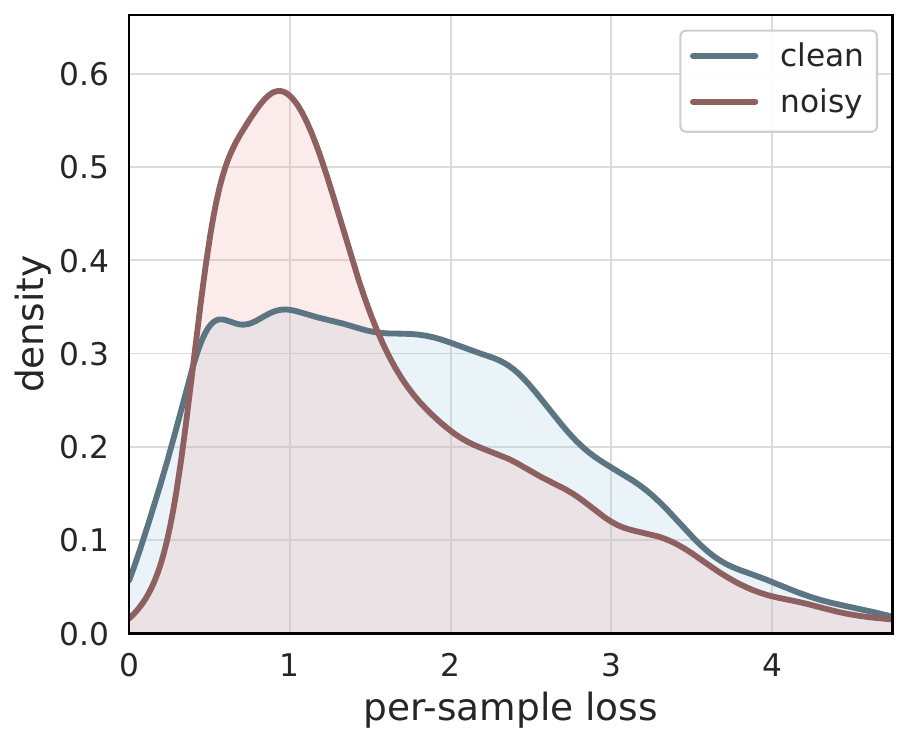}
  \caption{\ctype{contrast} (5)}
\end{subfigure}\hfill
\begin{subfigure}[t]{0.32\columnwidth}
  \centering
  \includegraphics[width=\linewidth]{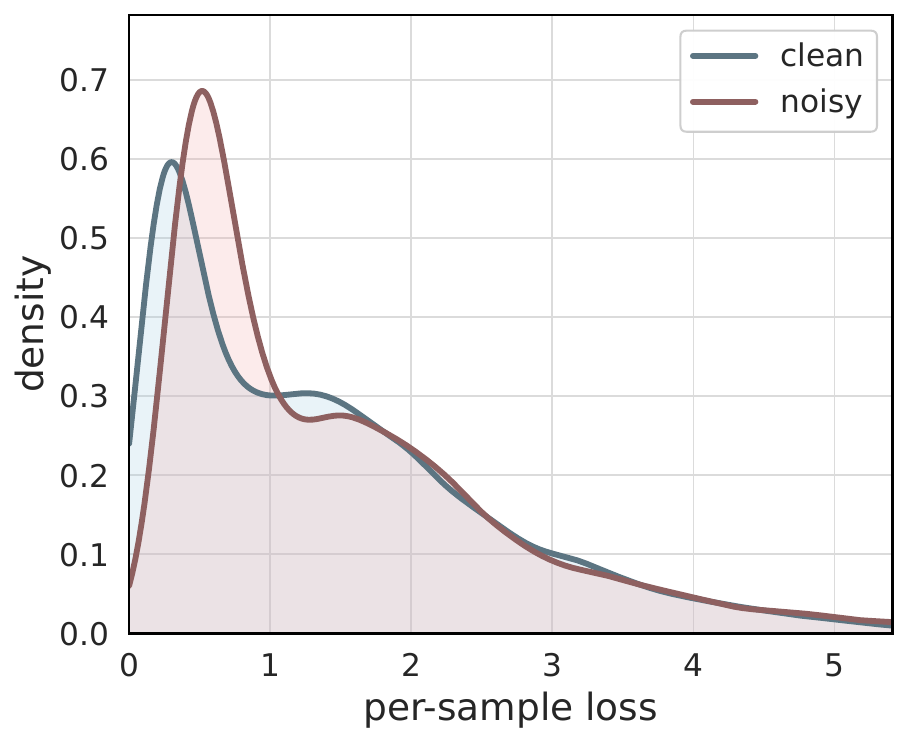}
  \caption{\ctype{glass} (3)}
\end{subfigure}

\begin{subfigure}[t]{0.32\columnwidth}
  \centering
  \includegraphics[width=\linewidth]{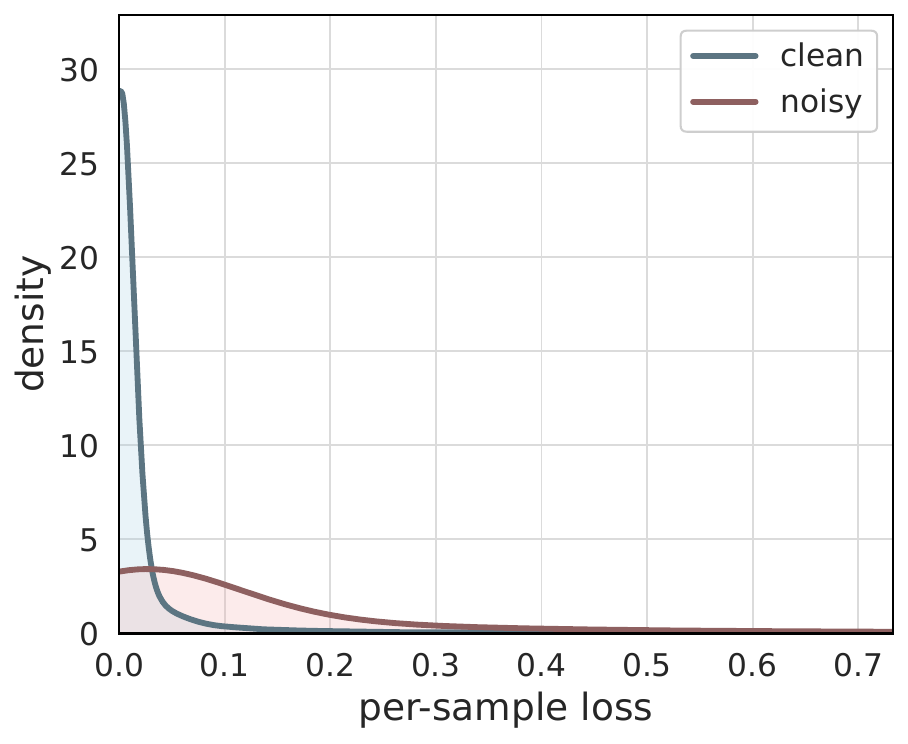}
  \caption{\ctype{frost} (1)}
\end{subfigure}\hfill
\begin{subfigure}[t]{0.32\columnwidth}
  \centering
  \includegraphics[width=\linewidth]{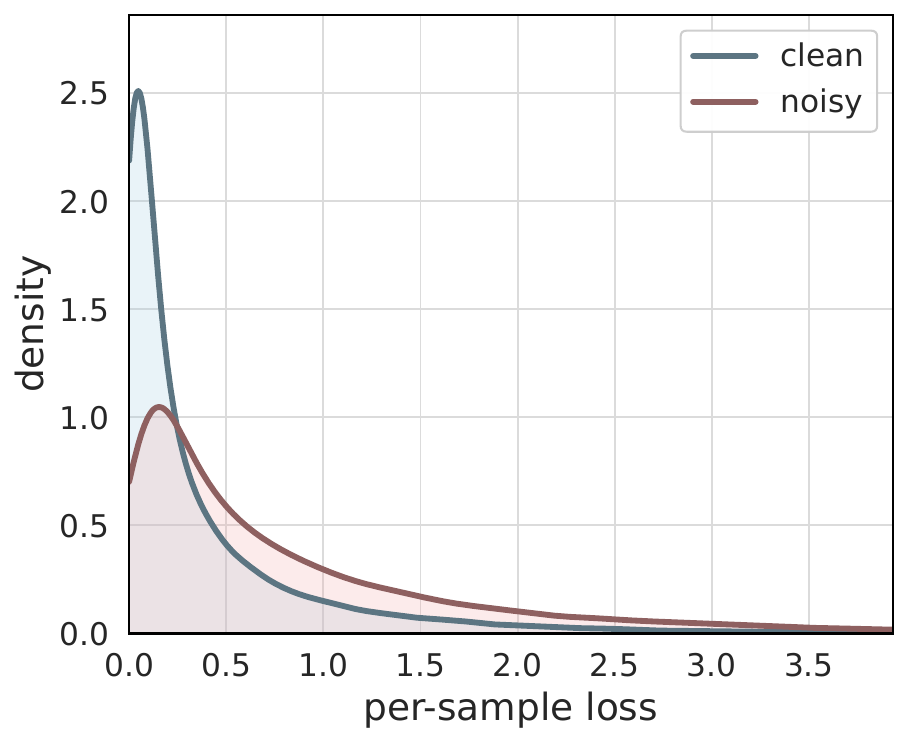}
  \caption{\ctype{contrast} (5)}
\end{subfigure}\hfill
\begin{subfigure}[t]{0.32\columnwidth}
  \centering
  \includegraphics[width=\linewidth]{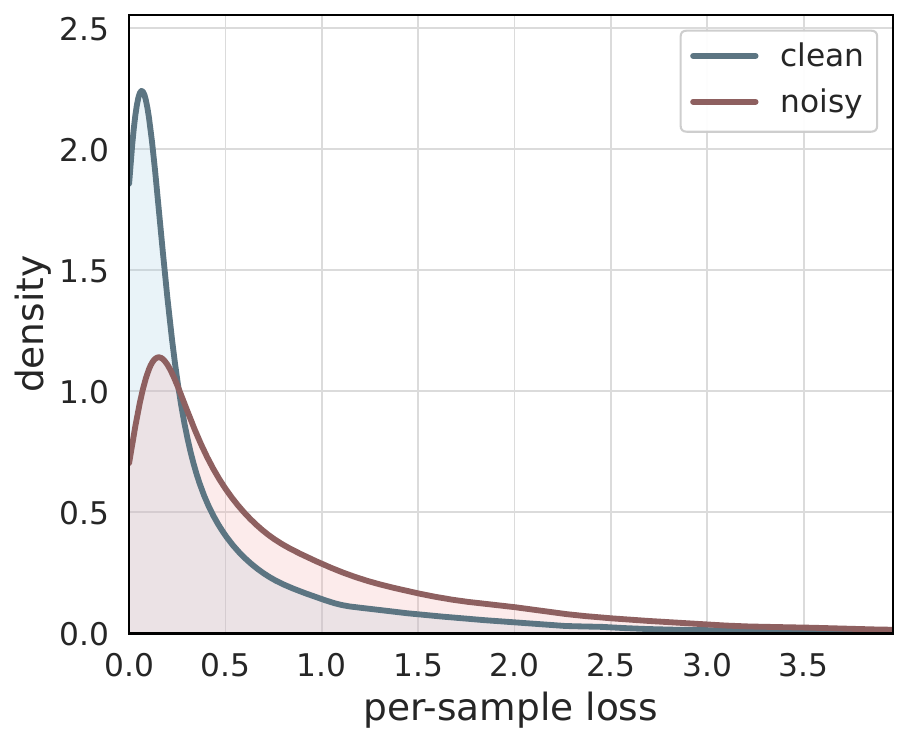}
  \caption{\ctype{glass} (3)}
\end{subfigure}

\caption{Per-sample loss distributions for clean and noisy samples under three representative \cilnbench{} settings. The top row shows the distributions at the end of the warmup stage, while the bottom row shows the distributions at the final training epoch.}
\label{fig:dm_bimodality}
\end{figure}

\subsubsection{Downstream Learning Behavior}
\label{sec:downstream}

We evaluate whether different benchmark generation mechanisms lead to different conclusions about noise-robust learning. We compare ERM, Co-Teaching~\cite{han2018coteaching}, and DivideMix~\cite{li2020dividemix} on \plidn{} and \cilnbench{} at comparable noise rates. For \cilnbench{}, we consider two training scenarios: \emph{clean-img}, where learners are trained on clean images paired with corruption-induced labels, and \emph{noisy-img}, where learners are trained on the corresponding corrupted images and labels. Results are reported as clean-test accuracy averaged over three random seeds.

Table~\ref{tab:cifar_overview} shows that \plidn{} and \cilnbench{} produce similar downstream behavior at low and medium noise rates. Across these settings, ERM, Co-Teaching, and DivideMix achieve comparable performance on both benchmarks. The differences become apparent at high noise rates. While Co-Teaching and DivideMix remain reasonably effective on \plidnH{}, both methods experience substantial performance degradation on several high-noise \cilnbench{} settings. In contrast, ERM remains comparatively stable.

The results suggest that benchmark behavior depends not only on the amount of label noise, but also on its structure. In \plidn{}, noisy labels originate from imperfect raters and are largely independent of the input seen by the learner. In \cilnbench{}, noisy labels are induced by corruptions that alter the input itself. As a result, some incorrect labels become consistent with the corrupted image and are therefore easier for the learner to fit. This weakens the small-loss assumption underlying methods such as Co-Teaching and DivideMix, which rely on noisy examples producing larger training losses than clean examples.

Table~\ref{tab:downstream_diagnostics} provides evidence for this effect. Several high-noise \cilnbench{} settings exhibit attractor classes that collect examples from many different true classes. These concentrated error patterns reduce class purity and can make noisy examples appear easier to learn than clean examples. Consequently, methods that identify noisy samples using training loss may incorrectly retain mislabeled instances while discarding informative clean ones.

We further evaluate the clean-start variant, \cilnbenchC{}, on representative high-noise settings. Although \cilnbenchC{} removes examples that already exhibit voter disagreement before corruption and therefore produces slightly lower noise rates, the same qualitative behavior remains. Co-Teaching continues to perform poorly, while ERM remains relatively stable. This indicates that the observed failures are not caused by pre-existing ambiguity in the data, but by the structure of the corruption-induced label noise itself.

Overall, the results show that methods that perform well on existing IDN benchmarks do not necessarily generalize to corruption-mediated label noise. At low and medium noise rates, \cilnbench{} behaves similarly to \plidn{}. At high noise rates, however, \cilnbench{} exposes failure modes that are not apparent under conventional benchmark generation procedures, highlighting the importance of evaluating noise-robust learning methods under multiple noise-generation mechanisms.

\subsection{Discussion and Takeaways}
\label{sec:discussion}

The results show that \cilnbench{} provides a mechanism-controlled way to generate instance-dependent label noise. Unlike rater-based benchmarks, where the source of uncertainty is implicit in the rater pool, \cilnbench{} makes the ambiguity mechanism explicit through the corruption type and severity. The benchmark covers $90$ settings across \cifar{}, \mnist{}, and \adult{}, and each setting includes the clean input, corrupted input, ground-truth label, aggregated voter distribution, individual voter distributions, and corruption metadata. This allows users to work with soft labels, sampled hard labels, or argmax hard labels under the same construction.

A key takeaway is that corruption-induced IDN is structured. Corruption type controls the direction of the distributional shift, often producing attractor classes, while severity controls the strength of the shift. This structure is not arbitrary label flipping: it reflects how a specific corruption interacts with the data. The positive \nth{} values across settings show that the resulting noise is instance-dependent, and the variation across corruption families shows that this dependence can be controlled.

The downstream results show that noise rate alone is not enough to characterize benchmark difficulty. At low and medium noise rates, \cilnbench{} and \plidn{} lead to similar conclusions for ERM, Co-Teaching, and DivideMix. At high noise rates, however, several \cilnbench{} settings expose failure modes that are not visible on \plidn{}. In particular, attractor-heavy corruptions can make wrong labels consistent with the corrupted inputs, so mislabeled examples may no longer have high loss. This weakens the small-loss signal used by Co-Teaching and DivideMix and explains why these methods can fail even when the nominal noise rate is comparable to \plidnH{}.

The clean-start results strengthen this interpretation. \cilnbenchC{} removes examples that already exhibit voter disagreement before corruption, so the remaining label errors are more directly attributable to the corruption mechanism. The same qualitative failures still appear under \cilnbenchC{}, indicating that the downstream difficulty comes from the structure of corruption-induced noise rather than from pre-existing ambiguity in the original data.

These findings suggest several uses for \cilnbench{}. It can be used to test whether noise-robust methods handle attractor-class effects, to study how performance changes as severity increases within a fixed corruption family, and to compare robustness across image, digit, and tabular settings under a shared benchmark construction. More broadly, \cilnbench{} complements rater-based IDN benchmarks by providing a controlled way to study how observable data degradation creates label uncertainty.

The benchmark also has limitations. The induced noise depends on the chosen voter pool, so different voters may shift which classes become attractors. The downstream learning experiments are currently limited to \cifar{}, and extending them to \mnist{} and \adult{} would help determine whether the same failure modes appear in other modalities. Finally, \cifarh{} is used as a reference for human annotation uncertainty, but it should not be interpreted as ground truth for how humans would label severely corrupted images.
\section{Related Work}

This work relates to benchmark construction for noisy-label learning, corruption-based robustness evaluation, and learning under noisy labels. Our framework connects the first two directions by using controlled input corruptions to generate realistic and interpretable instance-dependent label noise.

\subsection{Benchmark Construction for Label Noise}

A central challenge in noisy-label research is constructing benchmarks that balance realism and experimental control.

\subsubsection{Symmetric and Random Label Noise}

The most common synthetic benchmark strategy introduces label noise through random label flipping. Under symmetric noise, each label is independently replaced with an incorrect label according to a fixed probability \cite{natarajan2013learning}. These benchmarks provide precise control over noise rates and are widely used for evaluation, but they fail to capture realistic annotation behavior because all examples are equally likely to be mislabeled regardless of their difficulty.

\subsubsection{Class-Conditional Noise}

Class-conditional noise (CCN) allows corruption probabilities to depend on the true class \cite{patrini2017loss}. Label flips are governed by a class-transition matrix, enabling semantically plausible mistakes such as confusing cats with dogs more often than cats with airplanes. Although more realistic than random flipping, CCN still assumes that all examples within a class share the same corruption process and therefore ignores instance-specific ambiguity.

\subsubsection{Instance-Dependent Noise}

Instance-dependent label noise (IDN) models settings in which the probability of mislabeling depends on the input instance itself \cite{xia2020partdependent,berthon2021confidence}. This more closely reflects practical annotation processes, where difficult examples are naturally more likely to receive incorrect labels.

Several synthetic IDN generation methods have been proposed. Xia et al.~\cite{xia2020partdependent} assign each example an individual corruption probability using feature-space transformations. Gu et al.~\cite{gu2022idn} generate IDN through disagreement among a pool of classifier models acting as synthetic annotators. By varying model architectures and training conditions, their framework produces controllable noise levels and label distributions that resemble human uncertainty datasets such as CIFAR10-H \cite{peterson2019cifar10h}.

Despite their differences, existing IDN benchmarks primarily generate ambiguity through latent feature transformations, instance difficulty, or annotator variation. Consequently, the source of ambiguity remains implicit. In contrast, our framework generates ambiguity through controlled data corruptions, providing explicit control over both the source and severity of the induced uncertainty.

Several datasets provide naturally occurring noisy labels, including Clothing1M \cite{xiao2015learning}, Food101-N \cite{lee2018clean}, WebVision \cite{li2017webvision}, and CIFAR10-H \cite{peterson2019cifar10h}. While these datasets capture realistic annotation uncertainty, they offer limited control over the underlying noise-generation process.

\subsection{Corruption-Based Robustness Benchmarks}

Corruption benchmarks evaluate model robustness under degraded inputs. Representative examples include CIFAR-C and ImageNet-C \cite{hendrycks2019corruptions}, which apply common corruptions such as blur, noise, weather effects, compression artifacts, and geometric distortions across multiple severity levels. Subsequent work has extended this paradigm to broader forms of distribution shift and robustness evaluation \cite{mu2021validity}.

Our goal differs: rather than evaluating prediction accuracy under corruption (the standard use of CIFAR-10-C and ImageNet-C), we use corruptions as a mechanism for generating label uncertainty. The corrupted inputs are used during benchmark construction, while downstream learners can be trained on either clean or corrupted inputs paired with the generated labels. The objective is therefore benchmark construction rather than robustness evaluation.

\subsection{Learning under Noisy Labels}
\label{sec:rw_noisy_label_learning}

A large body of work studies learning from incorrect labels. Existing methods are commonly grouped into sample-selection, label-correction, and hybrid approaches. Sample-selection methods exploit the observation that neural networks fit clean examples faster than noisy ones \cite{arpit2017memorization}. Co-Teaching \cite{han2018coteaching} trains two networks that exchange small-loss samples, reducing confirmation bias while relying on the assumption that clean examples occupy the low-loss region. Label-correction methods attempt to repair suspicious labels rather than discard them. Examples include transition-matrix methods \cite{patrini2017loss}, Bootstrapping \cite{reed2015training}, and Confident Learning \cite{northcutt2021confident}, which relabel or remove examples based on model predictions and estimated noise patterns. Hybrid approaches combine both ideas. DivideMix \cite{li2020dividemix} fits a two-component Gaussian Mixture Model (GMM) to the loss distribution to separate likely clean and noisy samples. The noisy subset is then treated as unlabeled and incorporated into a MixMatch-style semi-supervised pipeline \cite{berthelot2019mixmatch}.

Both Co-Teaching and DivideMix depend on noisy examples being harder to fit than clean ones, creating a bimodal loss distribution. This assumption generally holds in conventional synthetic benchmarks. However, Section~\ref{sec:downstream} shows that corruption-mediated noise can violate it: when an incorrect label remains plausible for a corrupted input, noisy examples become as easy to fit as clean ones, causing the bimodal structure to collapse. Consequently, both methods fail on the same high-noise settings for the same underlying reason. Since noisy-label methods are often evaluated on synthetic benchmarks, benchmark design can strongly affect conclusions about algorithm performance. Our work provides a complementary benchmark-generation framework for studying how different sources of ambiguity influence noisy-label learning.

\section{Conclusion and Future Work} \label{sec:conclusion}

We introduced \cilnbench{}, a benchmark-generation framework for instance-dependent label noise based on controlled input corruptions. By making the source and severity of ambiguity explicit, \cilnbench{} produces controllable instance-dependent noise across image and tabular datasets. Our experiments show that the generated benchmarks exhibit genuine instance dependence, can produce label distributions closer to human uncertainty than existing synthetic IDN benchmarks, and expose failure modes of noise-robust learning methods that remain hidden under traditional rater-fallibility benchmarks. Overall, our results suggest that benchmark difficulty depends not only on the amount of label noise, but also on its underlying structure, making corruption-mediated and rater-mediated IDN complementary evaluation settings.

Several directions remain for future work. First, the downstream evaluation could be extended beyond CIFAR-10 to verify whether the observed failure modes generalize to other modalities. Second, the benchmark could be expanded with additional corruption families and domain-specific degradations. Finally, the benchmark opens opportunities for developing mechanism-aware noisy-label learning methods that explicitly exploit information about the source and structure of ambiguity rather than relying solely on loss-based heuristics.

\section{Acknowledgments}

\paragraph*{AI-Generated Content Acknowledgment}
Claude Opus 4.7 was used for code generation from author-provided specifications and GPT 5.2 was used for language editing. All generated content was reviewed and validated by the authors. The research design, experiments, analysis, conclusions, and all claims regarding prior work are the authors' own intellectual contributions.

\bibliographystyle{IEEEtranN}
\bibliography{ref}

\end{document}